\documentclass{ieeeaccess}

\usepackage{amssymb}
\usepackage{amsmath}
\usepackage[hyphens]{url}

\usepackage{xspace}
\usepackage{setspace}

\usepackage{booktabs}

\usepackage{algorithm, algorithmic}

\usepackage{cite}
\usepackage{,amsfonts}
\usepackage{graphicx}
\usepackage{caption}
\usepackage{textcomp}
\def\BibTeX{{\rm B\kern-.05em{\sc i\kern-.025em b}\kern-.08em
    T\kern-.1667em\lower.7ex\hbox{E}\kern-.125emX}}

\begin{document}

\history{Date of publication xxxx 00, 0000, date of current version xxxx 00, 0000.}
\doi{10.1109/XXX.2025.DOI}

\renewcommand{\thevol}{X}
\renewcommand{\theyear}{2025}

\title{Finding One's Bearings in the Hyperparameter Landscape of a Wide-Kernel Convolutional Fault Detector}

\author{
  \uppercase{Dan Hudson}\authorrefmark{1}, %\IEEEmembership{Fellow, IEEE},
  \uppercase{Jurgen van den Hoogen\authorrefmark{1}, and Martin Atzmueller}\authorrefmark{1,2},
%\IEEEmembership{Member, IEEE}
}
\address[1]{Semantic Information Systems Group, Osnabrück University, Wachsbleiche 27, 49 Osnabrück, Germany}
\address[2]{German Research Center for Artificial Intelligence (DFKI), Hamburger Straße 24, 49084 Osnabrück, Germany}

\tfootnote{This work was supported by funds of zukunft.niedersachsen, Volkswagen Foundation (project “HybrInt -- Hybrid Intelligence through Interpretable Artificial Intelligence in Machine Perception and Interaction”).}

\markboth
{D. Hudson \headeretal: Finding One's Bearings in the Hyperparameter Landscape of a Wide-Kernel Convolutional Fault Detector}
{D. Hudson \headeretal: Finding One's Bearings in the Hyperparameter Landscape of a Wide-Kernel Convolutional Fault Detector}

\corresp{Corresponding author: Dan Hudson (e-mail: daniel.dominic.hudson@uos.de).}

%% Abstract
\begin{abstract}

State-of-the-art algorithms are reported to be almost perfect at distinguishing the vibrations arising from healthy and damaged machine bearings, according to benchmark datasets at least.
However, what about their application to new data?
In this paper, we confirm that neural networks for bearing fault detection can be crippled by incorrect hyperparameterisation, and also that the correct hyperparameter settings can change when transitioning to new data. 
The paper combines multiple methods to \emph{explain} the behaviour of the hyperparameters of a wide-kernel convolutional neural network and how to set them. 
Since guidance already exists for generic hyperparameters like minibatch size, we focus on how to set
architecture-specific hyperparameters such as the width of the convolutional kernels, a topic which might otherwise be obscure. 
We reflect different data properties by fusing information from seven different benchmark datasets, and our results show that the kernel size in the first layer in particular is sensitive to changes in the data. 
Looking deeper, we use manipulated copies of one dataset in an attempt to spot why the kernel size sometimes needs to change. The relevance of sampling rate is studied by using different levels of resampling, and spectral content is studied by increasingly filtering out high frequencies. We find that, contrary to speculation in earlier work, high-frequency noise is not the main reason why a wide kernel is preferable to a narrow kernel. 
Finally, we conclude by stating clear guidance on how to set the hyperparameters of our neural network architecture to work effectively on new data.

\end{abstract}

%% Keywords
\begin{keywords}
Bearing Fault Detection, CNN, Deep Learning, Industrial Fault Detection, Hyperparameter Analysis, Time Series Classification



\end{keywords}

\titlepgskip=-15pt

\maketitle

\section{Introduction}
\label{sec:intro}

Machines are full of bearings, small metal balls usually arranged in a circular casing, which roll around and allow one piece of a machine to rotate relative to another. 
Even during their ordinary working life it is possible for bearing components to pick up damage such as, e.g., a tiny scratch in the casing. 
As the ball bearings roll over that scratch again and again (perhaps hundreds of times per minute for days, weeks and months), it can grow into a structural weakness. One day, the casing can crack due to this weakness, breaking the rotating element and putting the entire machine out of commission until repairs are made. Therefore, methods for detecting such faults are crucial.
Unfortunately the accumulating damage cannot be seen from outside the machine, and so practitioners often turn to recorded vibrations which spread easily from the inside to the outside. 

At the time of writing, the most effective method for analysing vibration signals in order to identify faulty bearings is to use neural networks. Neural networks have demonstrated excellent accuracy at classifying different types of bearing vibrations but their high accuracy is not guaranteed. 
Even the best-performing neural networks have multiple `hyperparameters' -- decisions that are left to the user yet which can determine whether the network succeeds or not. We contend that although hyperparameters are crucial they are under-researched within the field of bearing fault research. 
Throughout the present paper we take a recent state-of-the-art neural network and ask how its performance changes as a result of changing its hyperparameters. By the end of the paper we can sketch a map of the ``hyperparameter landscape'' of our chosen network, showing how the neural network responds as we traverse combinations of hyperparameters. 

This paper is a significantly extended and adapted revision of~\cite{hudson2024stay} building on our work presented in~\cite{hoogen2023hyperparams}. We present a series of new experiments that address questions raised and left unresolved by our earlier work in~\cite{hudson2024stay} and \cite{hoogen2023hyperparams}. Two new neural network architectures are studied in order to show how important hyperparameter configuration is to a wider range of networks for bearing fault detection. Presenting a novel experimentation methodology, extending the methodology presented in~\cite{hoogen2023hyperparams}, we run new experiments on 
manipulated subsets of the data to see how two main dataset properties (sampling rate and spectral content) affect the hyperparameter landscape. 
Overall, we present more datasets, more neural network architectures and more types of experiments compared to
\cite{hoogen2023hyperparams}. In the next subsections of this introduction, we briefly lay out the rationale of our work.

% \subsection{notes from lit review}

% \cite{sharma2019hyperparameter}
% lodge a similar argument to ours, proposing that previous work focuses too much on which algorithm is best on a narrow set of testing data, rather than on building knowledge that generalises across many datasets. So, they choose to focus on what hyperparameters are important. 
% States that for new algorithms it is usually unknown which hyperparameters to tune and what ranges of values are reasonable to sample from. 

% \cite{theodorakopoulos2024hyperparameter}
% notes the importance of hyperparameter configuration for accurate modelling and prediction. Notes that insights are gained from designing better configuration spaces and computing hyperparameter importance scores. 

\subsection{Finding Faults and Beating Benchmarks}

In the pursuit of effective algorithms to recognise when damage has occurred to a bearing element, researchers have settled on a series of `benchmark' datasets, used to train and test different approaches. 
%The obvious benefit of these openly-available datasets is that it is like having a fair race: everyone is starting at the same place and running the same route, so we can compare algorithms fairly based on their performance on these benchmark datasets. 
These openly-available datasets act like a fair contest through which we can compare the accuracy of different algorithms. 
Operating on well-known benchmark data, various types of neural networks have been reported to achieve 97\%+ accuracy on
fault detection tasks, in papers such as \cite{zhang2018deep, eren2017bearing, guo2016hierarchical, han2021combination}.  A recent review of the literature on the topic \cite{neupane2024data} indicates that at least 25 different deep learning algorithms have achieved over 95\% accuracy (often 99\% or 100\%) on one or more bearing fault benchmarks. 

Such consistently high levels of performance imply that the difference between two state-of-the-art algorithms is now predestined to be small and there is little room left to improve beyond previous results. Therefore, comparing one state-of-the-art option to another is becoming less interesting and important than before. 
What we find more interesting is the question of whether they will still achieve near-perfect results when someone applies them in real life, when the sensors and machines are different from what the benchmarks used. 

%Speaking more generally about neural networks, when a practitioner takes an existing algorithm and tries using it on their own data, there is a risk that it does not work. Imagine that the practitioner is left in this situation. What can they do? 
Imagine that a practitioner tries an existing algorithm  on their own data and it does not work. What can they do? 
One of the first things they can try is to modify the so-called `hyperparameters', decisions about how to structure and train the network. %, 
%such as choosing the minibatch size (the number of examples used in a single step of training). 
% These hyperparameters control how to generate the training signal that the neural network learns from -- they are used in generating the numbers that incrementally update the neural network.  
% Just like how a mis-tuned radio will not play your favourite radio show, it is possible to mis-tune hyperparameters and end up with a training signal that does not do what you want, i.e., it will not teach the neural network how to do the task. 
It has been noted that the performance of most machine learning algorithms depends a lot on hyperparameters. Tuning hyperparameters to their best values can be at least as important as the choice of algorithm \cite{pfisterer2021learning} or chasing after marginal gains on the state-of-the-art by developing new architectures \cite{wistuba2015hyperparameter}. 
%
% Alongside hyperparameters that modulate the training process, one can consider  `architectural hyperparameters' which control the size and shape of a neural network. 
% Returning specifically to bearing fault detection, we noted earlier that the differences appear to be minimal between the top 25 neural networks for detecting damaged bearings. By contrast, larger differences in accuracy can emerge when the hyperparameters of a network are changed. 
For example, \cite{kim2022effect} reports that the accuracy of a bearing fault classifying convolutional neural network is cut in half simply as a result of increasing from 6 convolutional layers to 8. 
When it comes to getting near-perfect accuracy on bearing fault detection, getting the hyperparameters right might be more important than choosing between neural network architectures.

\subsection{Shaping the Neural Network}

Architectural hyperparameters like the number of layers in a network affect how complicated it is to process each input. Making the network smaller reduces the number of internal parameters it has, meaning that the computer needs to perform fewer arithmetic operations. This has a knock-on effect %that impacts, for example, how long it takes to train the network. S
meaning that smaller networks train faster and are quicker to generate predictions on new data. If some architectural hyperparameters can be reduced without harming the accuracy, it would be good to know this. Practitioners would benefit from knowing that they can make the network smaller and thus faster to train and use. Moreover, in real-world settings, there is the possibility of `embedding' fault detection algorithms, so that they run on small devices situated in and amongst the machines they analyse rather than running on a user's laptop. This would require networks that can run using reduced storage, memory and processor resources.  %Embedded processing skips the step of extracting data from the sensors and communicating it to a central location (like a laptop), and removes the need to keep a laptop or computer running continuously in order to process the data. An additional benefit of small networks is that they can be run on smaller and cheaper embedded devices. Clearly, there is 
%real value to understanding the hyperparameters of a network, to see which hyperparameters can and cannot be changed when processing a given dataset. In this vein, we have worked to understand the architectural hyperparameters of a network capable of state-of-the-art results on fault detection tasks. 

Of course, there must be limits to how much we can shrink a network or otherwise manipulate its hyperparameters. Some changes will be too drastic to work and they will make the neural network less able to detect bearing faults. 
% One line of our research is to study which hyperparameters are crucial and require the most careful handling. %We wish to understand how much impact different changes will have and be able to state the relative importance of the hyperparameters. 
% Re-examining the wide-kernel network presented in ~\cite{zhang2017new,hoogen2020improved}, we distil new explanations of how the network's hyperparameters impact its accuracy at classifying bearing fault signals. Looking at other types of neural network, previous research has already indicted that not all hyperparameters are equal and it is reasonable to expect that at least some of the hyperparameters can be modified without disastrous consequences for the accuracy of the neural network \cite{bergstra2012random}. 
%
How many layers is enough in a neural network? Why do we set other hyperparameters in the way that we do? 
Picking two fault detection papers to use as an example, we find that \cite{eren2017bearing} uses 20 hidden units in fully-connected layers whilst \cite{8756423} uses 64 instead. As is typical in the field there is no stated reason for why these particular numbers should be used, resulting in the danger that if future users ever need to adapt these hyperparameters for new datasets, they could justifiably feel lost. We suspect that what users in fact want is not just an algorithm that performs well on benchmarks, but also a simple process for setting the hyperparameters when moving to new data. We use an approach called ``multiple defaults'' to very efficiently tune the hyperparameters of a neural network for bearing fault classification.

\subsection{Do Dataset Properties Determine Performance?}

Datasets have different properties, providing one reason why hyperparameters might need (re)tuning on new data even though they were already working on benchmarks. %In the field of bearing vibration, research focuses on sensor recordings stored in `time series', so called because they consist of a series of sequential measurements of some variable over time. 
To record vibrations as time series leads to them having several important properties, including the sampling rate which determines how many measurements are made per second and conversely how far apart in time each measurement point is. Sampling at 12kHz, the movement of a vibrating machine is measured every 0.00008333 seconds. 
The sampling rate affects the dataset in multiple ways. 
First, an increased sampling rate means that more numbers are used to record a vibrating machine, implying that the neural network needs to do more computation in order to process the inputs. Likewise it is plausible that the size of `convolutional filters', which act like stored patterns in certain neural networks, needs to be bigger when dealing with a higher sampling rate.

The sampling rate additionally affects what frequency bands can be detected, with half the sampling rate (also known as the `Nyquist frequency') providing the upper limit of what frequencies can be captured. 
One open question is how much the highest frequencies in vibration recordings are used by neural networks. This is interesting because 
if the highest frequencies are not needed, then storing and processing the recordings could be made more efficient by reducing the sampling rate. 
In fact, if high frequencies are unnecessary for fault detection then removing them could be a form of noise filtering. 
This is not implausible, since many signals resulting from vibrations (like from musical instruments) have an overtone series, implying that there is redundancy. 
As noted in \cite{lacey2008vibration}, vibrations in bearing elements are transmitted to the sensors through the rest of the machine, which also consists of multiple components that themselves vibrate. This potentially leads to irrelevant or misleading information in the recorded signal, and if some frequency bands are just noise then it could be beneficial to remove them. 
Our research looks at how a wide-kernel convolutional network responds to changes in the sampling frequency. To further pinpoint the matter, we also ask whether filtering the irrelevant frequencies without changing the sampling rate is helpful. This leads us to manipulate one of the benchmark datasets in two ways: resampling, and, filtering.

\subsection{The Paper Ahead}

The paper ahead re-examines and then extends upon the results from several pieces of our own past work.
In addition to providing new insights on past results, this present study presents new experiments with resampling and filtering of the data, allowing us to investigate how the properties of the data affect hyperparameter tuning.

We use a cocktail of multiple methods to explain the impact of hyperparameters on a wide-kernel CNN's accuracy. We summarise results from past work in order to see the average performance level when specific hyperparameter values are carried across seven datasets. We also gather new results from experiments on manipulated versions of the data. %We quantify the `feature importance' -- the extent to which different variables contribute to an outcome -- of different hyperparameters and how it changes when processing either resampled or filtered copies of a dataset.  
This lets us find out how relevant dataset properties like the sampling rate are to deciding what hyperparameters to use. 
By specifying multiple defaults -- a sequence of hyperparameter settings to try out on new data -- we give users robust guidance on how to tune our chosen wide-kernel network architecture on new data.
We also undertake separate additional experiments with two other architectures (besides the wide-kernel CNN) in order to round out the generality of our basic premise. 

The key contributions of this work are:
\begin{itemize}
    \item We show that hyperparameters are important across multiple types of neural networks rather than a single network type. Past results and new experiments with additional neural network architectures confirm this.  
    %\item We show that hyperparameters affect how accurate a neural network is (the majority of our work specifically investigates a wide-kernel CNN architecture). 
    %\item Fusing information from 7 datasets, we show that hyperparameters that are successful on one dataset are not necessarily successful on another dataset. 
    % \item We also highlight that not all hyperparameters are equally important.
    \item We propose that properties of the dataset, such as sampling rate, impact the optimal values of hyperparameters, which leads us to run experiments with manipulated versions of the time series data. 
    \item After integrating the information from multiple different experiments, we find that due to redundancy in the signal, very high frequencies are not needed for successful bearing fault detection.
    \item We give a precise process, based on the notion of ``multiple defaults'', for how to set the hyperparameters of a wide-kernel CNN when transitioning to new data. 
\end{itemize}

\section{Background \& Related Work}

In this section, we review fault detection for rotating machines in industrial settings. We also provide an overview of deep learning (DL) in the realm of time series analysis using sensor data, emphasising the design and training of convolutional neural networks (CNN). Additionally, given that this research concentrates on the architectural hyperparameters of a neural network, we discuss studies related to hyperparameter optimisation.

\subsection{Fault Detection}

%Detecting faults in rotating industrial machinery is essential to avoid breakdowns~\cite{motor1985report}. Fault detection data are typically collected from sensors that measure vibrations, represented as time series data. 
Initially, fault detection relied on physics-based models that required an understanding of the underlying mechanisms that generate and propagate vibrations~\cite{yin2014data}. However, these models struggled to adapt to changing environments and increasing data complexity. 
The advent of the Industrial Internet of Things (IIoT) and data-driven analysis techniques has revolutionised fault detection methods, enabling a more intelligent and automated approach~\cite{8756423,zhao2019deep:b}. These advances eliminate the need for in-depth technical knowledge of industrial machinery, allowing for automated processing and adaptation to changing operational environments. For example, machine learning approaches such as 
K-NN~\cite{pandya2013fault,6967842}, Random Forest~\cite{7974734}, and SVM~\cite{santos2015svm,you2014wpd,huang2011support,konar2011bearing} have been applied for fault detection. However, these methods require extensive feature extraction, a time-consuming process that has to be fine-tuned towards the type of data used. 
With the development of deep learning techniques, these steps were no longer necessary since deep learning techniques are able to automatically extract features from raw data, simplifying the otherwise complex feature extraction process. 

Multilayer perceptrons (MLPs), also known as feedforward networks, are a relatively simple form of deep learning that have been applied to various domains of time series analysis, including stock prediction~\cite{naeini2010stock}, weather forecasting~\cite{abhishek2012weather}, and fault detection~\cite{hajnayeb2011application}. Subsequently, recurrent neural networks (RNNs)~\cite{malhotra2016lstm,hannun2014deep} and convolutional neural networks (CNNs)~\cite{zhang2017rolling,zhao2019deep:b,s19235300,8756423,yang2015deep} demonstrated significant performance improvements in time series tasks. %CNNs, particularly one-dimensional (1D) CNNs, were designed to leverage automated feature extraction and process raw time series data effectively~\cite{ince2016real,zheng2014time}. 

CNN methods, when combined with data transformations like spectrograms, have been employed multiple times~\cite{zhao2019deep:a,chen2019intelligent}.
However, 1D CNNs are able to process the raw time series data directly (without the need for spectrograms) and integrate automated feature extraction, and are often used in fault detection~\cite{ince2016real,zheng2014time}. These 1D CNNs also tend to be resilient to noise in time series data and can be trained with a small sample size~\cite{zhang2019limited}.
Over time, numerous refinements for fault detection using convolutions have been introduced. These include, but are not limited to, wide-kernel CNNs~\cite{zhang2017new,hoogen2020improvedWDCNN,hoogen2021} (which are used in this work), attention CNNs~\cite{zhao2023fault}, periodic CNNs~\cite{periodnet}, and CNN autoencoders for unsupervised learning tasks~\cite{CHEN202054}. Each of these networks has unique characteristics regarding complexity, computational requirements, and specific tasks.
%The general mechanics behind CNNs are further discussed in the next paragraph.

% \subsubsection{Convolutional Neural Networks} 
% Convolutional Neural Networks (CNNs) are a specialised type of neural network that were originally created to handle two-dimensional (2D) data, such as images. Introduced by LeCun in 1989~\cite{lecun1998gradient,lecun1995convolutional}, CNNs utilise a feed-forward architecture that applies convolutions instead of general matrix multiplications, making them highly effective for data organised in a grid-like structure. In contrast to conventional MLPs, CNNs have become essential in areas such as Computer Vision, Image Recognition~\cite{lecun1995convolutional,simonyan2014very}, and are increasingly used in time series analysis.

The primary benefits of CNNs compared to other neural networks arise from their use of local receptive fields, weight-sharing, and sub-sampling~\cite{goodfellow2016deep}. These characteristics greatly decrease memory usage and computational complexity, thereby improving algorithmic efficiency.
Convolutional layers apply filters to %convolve 
input data and this basically allows the network to identify where patterns occur within a large signal. 
The activation function, commonly a Rectified Linear Unit (ReLU), introduces non-linearity, allowing the network to learn intricate patterns~\cite{gulcehre2016noisy}. After convolution and activation, a pooling layer often shrinks the signal's size in order to decrease the number of parameters of the next layer which helps prevent overfitting and lowers computational demands.

% Initially, CNNs were less common for time series data due to their 2D nature, requiring conversion of one-dimensional (1D) data into 2D matrices, adding computational overhead. This changed with the development of 1D CNNs, which can directly process raw time series data~\cite{kiranyaz2015convolutional,hoogen2021,zhang2017new,zhang2019limited,hoogen2020improvedWDCNN,ince2016real}. These advancements have solidified CNNs as a powerful tool for time series analysis, capable of efficiently handling high-frequency sensor data and extracting valuable features with minimal preprocessing.

The convolution operation is mathematically expressed as:

\begin{equation}
y_{i}^{l+1}(j)=k_{i}^{l} * M^{l}(j)+b_{i}^{l},
\end{equation}

where $b_{i}^{l}$ is the bias, $k_{i}^{l}$ represents the filter weights, and $M^{l}(j)$ is the local region of the input in layer $l$. The weights of the filter act in some ways like a learned pattern, in the sense that the filter reacts most strongly (gives the largest numbers as outputs) when the input is large \textit{and} correlates with the weights of the filter. The filter applies to little regions of the overall signal with each region being one stride along from the previous one. In the notation above $i$ is a number used to refer to different regions.  
% The most common activation function is ReLU~\cite{agarap2019deeplearningusingrectified}, which introduces nonlinearity to the network. 
% Next, a pooling layer is utilised that allows the CNN
% to downsample the output of the convolutional layer 
% making it more resource efficient in 
% various applications from image and speech recognition to time series analysis.

\subsection{Hyperparameter Search}

Hyperparameter tuning appears most prominently in the field of ‘AutoML’, a research area based around the goal of finding and training the algorithm (which could be a neural network or another machine learning method) that will best perform a task, all without the involvement of humans.
Although the hyperparameter space is considered one of the main preoccupations of AutoML researchers \cite{baratchi2024automated}, it is often used simply to define which ranges of values will be considered ‘in scope’ when tuning an algorithm. 
%
%AutoML researchers are commonly preoccupied with three things: the hyperparameter space, the search strategy, and the evaluation method. 
%We focus on the first. 
%Hyperparameter (sub)spaces can be defined by deciding which ranges of values will be considered ‘in scope’ when tuning the hyperparameters of an algorithm. 
%A lot of research simply sees this as a preliminary step needed as an input into the hyperparameter search algorithm, which is treated as the main subject. 
However, %with a change of emphasis, it is possible to focus on 
better understanding the hyperparameter space %. Doing so 
gives us the chance to eventually build knowledge that generalises across many datasets instead of being restricted to concluding which algorithm is best on a narrow set of testing data \cite{sharma2019hyperparameter}. The hyperparameter space is a way of specifying all the values the hyperparameters of an algorithm can take. If we combine this with extra information about how well the algorithm performs when using each combination of hyperparameters, we get the so-called “response landscape” or “response surface”. 

% Hyperparameter tuning has considerable overlap with the field of ‘AutoML’, a research area based around the goal of finding and training the algorithm that will best perform a task, all without the involvement of humans. AutoML can be restricted to neural networks, but it does not need to be. It could even exclude neural networks as a possibility and only consider other machine learning algorithms like k-nearest neighbours or support vector machines. 

For example, the response landscapes of neural networks are crucial to the discussion of random search by Bergstra and Bengio \cite{bergstra2012random}. Their results suggest that many hyperparameters are not very important and thus it is not worth the time grid search (compared to random search) takes to try out each tuning option. This leads to the notion of “low effective dimensionality” in the response landscape, meaning that only a small number of changes need to be made to move from a low-performing set of hyperparameters to a high-performing set. With low effective dimensionality, random search is better than grid search. Additionally, the research concluded that different hyperparameters are important on different datasets. This implies dataset properties are an extra piece of the puzzle and the response landscape can change with the data. Bergstra and Bengio did not look further into %the links between 
dataset properties %and the response landscape since it was 
since doing so was not needed as part of their argument in favour of random search above grid search. However, we pick up on this line in our research, explicitly looking at the impact of the sampling rate and filtering of high frequencies within time series. 

%For example, one of the seminal works in AutoML is the study of Bergstra and Bengio \cite{bergstra2012random} to compare random search against grid search for hyperparameter optimisation. The authors compare two automated search strategies, and argue that random search has some properties that generally make it a better choice when tuning hyperparameters. What is most relevant for us is that they link their conclusions to the structure of the hyperparameter space. For example, the results suggest that many hyperparameters are not very important and thus it is not worth the time grid search takes to try out each tuning option. This leads to the notion of “low effective dimensionality” in the response landscape, meaning that only a small number of changes need to be made to move from a low-performing set of hyperparameters to a high-performing set. With low effective dimensionality, random search is better than grid search. It is worth noting however that the research also concluded that different hyperparameters are important on different datasets. This implies dataset properties are an extra piece of the puzzle and the response landscape can change with the data. Bergstra and Bengio did not look further into the links between dataset properties and the response landscape since it was not needed to argue in favour of random search above grid search. However, we pick up on this line in our research, explicitly looking at the impact of the sampling rate and filtering of high frequencies within time series. 

\subsubsection{Hyperparameter importance}

Quantifying “hyperparameter importance” is another relevant sub-strand of past work which aims to probe the relationships between hyperparameters and neural network performance. Hutter, Hoos and Leyton-Brown \cite{hutter2014efficient} established what is probably the most widely-used method to estimate how important hyperparameters are. % will be to the tuning of a machine learning algorithm. 
It uses a random forest as a ‘surrogate model’ predicting how well different hyperparameter combinations will perform. The surrogate model then leads to importance expressed as the variance explained (by the hyperparameter) within a functional ANOVA.  The functional ANOVA has various nice properties including that it states the variance explained by subsets of hyperparameters as well as by individual hyperparameters.
Moreover, the functional ANOVA makes it possible to ask ‘what performance is expected on average with \verb|Hyperparameter_A = value_X|’. Subsequent research re-used the idea of using an fANOVA to find hyperparameter importance. For example, later work applied the technique to residual neural networks for image classification \cite{sharma2019hyperparameter}, and multi-objective tasks where the goal is to find a Pareto front balancing network performance against considerations like training time and energy consumption \cite{theodorakopoulos2024hyperparameter}.
%The functional ANOVA also makes it possible to estimate marginals which are slightly different and ask what is average performance for particular settings of particular hyperparameters. So, they let the researcher ask ‘what performance would I expect on average with \verb|Hyperparameter_A = value_X|’. In the body of their paper, the authors introduce a new efficient method for computing marginals from random forest models. They substantiate their proposed method with a computational study based on WEKA’s hyperparameter space, which is built out of 768 hyperparameters, applied to several (now considered simple) ML baseline datasets. Subsequent research re-used the authors’ idea of using an fANOVA to find hyperparameter importance. Later work applied the technique to residual neural networks for image classification \cite{sharma2019hyperparameter}, and multi-objective tasks where the goal is to find a Pareto front balancing network performance against considerations like training time and energy consumption \cite{theodorakopoulos2024hyperparameter}, for example. 

The other common way to assess hyperparameter importance is to perform ablation studies. In order to do an ablation study it is necessary to have already tuned the algorithm, since the goal is to find a path from the fully-tuned algorithm back to the basic default values for each hyperparameter. The path is generated by a greedy algorithm which, at each step, selects the hyperparameter which 
drops performance the most when it is reverted back to its default.
%leads to the greatest performance drop when it is reverted back to its default. 
The final path passes through all the hyperparameters in the order in which they contributed the most to performance \cite{baratchi2024automated}. Compared to the fANOVA approach, an ablation study does not include information about regions of the hyperparameter space that are distant from the defaults and the tuned values. 

Compared to past work on hyperparameter importance, our own work has a number of differences worth pointing out, the first being that we focus on complex time series data analysed using modern neural network architectures. Time series datasets sophisticated enough to warrant neural networks are under-researched in connected to hyperparameter importance. We also use resampling and filtering to actively modify data rather than relying on pre-existing differences between datasets, because we place greater emphasis on understanding the context of bearing fault detection. % – our research is more applied than methodological. 
% Finally, we add in a complementary form of feature importance based on the Shapley values of a surrogate model. Although Shapley values have been mentioned in the context of Bayesian optimisation, as far as we are aware this has only been applied to other optimisation tasks, not hyperparameter optimisation \cite{adachi2023looping, rodemann2024explaining}. 

\subsubsection{Hyperparameter Search with Multiple Datasets}

Finding knowledge that can generalise across datasets is one of the goals of AutoML and numerous approaches have been proposed for how to do it. Most commonly, the default values of an algorithm offer a way to transfer knowledge between datasets. Bayesian optimisation using surrogate models is a widely-used method which allows for transfer across datasets by ‘warm-starting’ the search process \cite{wistuba2015hyperparameter, pfisterer2021learning}. Wistuba, Schilling and Schmidt-Thieme suggest a pruning method to leverage past experience from multiple datasets in order to speed up hyperparameter tuning %. They use Kendell’s tau to assess the similarity of the response landscape on a new target dataset to the landscapes of previously-explored datasets. This information is used to develop a surrogate model that can be used to guide the tuning of hyperparameters on new data 
\cite{wistuba2015hyperparameter}. 
Perrone et al. \cite{perrone2018scalable} trained a neural network to represent hyperparameter combinations in a way that makes predicting performance easier. The representations take hyperparameter values as inputs and transform them into 50-dimensional vectors 
that can be used as a basis for easily predicting how well the hyperparameters would do on a range of training datasets.
%, which in turn provide effective inputs to a linear regression that will predict how well the hyperparameters do on a range of training datasets. Using multiple datasets during training was intended to lead to vectors that generalise well to new data. 

Considering the aforementioned sources, it is clear that generalising across datasets has been considered in past literature. However, it has taken the form of using black-box methods to find a single set of optimal values. By contrast, we %aim to provide interpretable knowledge about the hyperparameters. We 
are interested in getting to know the hyperparameters across a multiple values rather than identifying a single optimal point (and ignoring the rest of the hyperparameter space).

Much hyperparameter optimisation research has relied on Bayesian optimisation applied to surrogate models as a way to explore the hyperparameter space and find effective settings to use on new data. An interesting alternative comes in the form of “multiple defaults”, which basically means a small list of tried-and-tested combinations. These combinations are handed to a new user along with an ML algorithm, letting the user know what are sensible combinations of hyperparameter settings to try out. Arguments in favour of this approach include how easy to implement and use it is, and that the hyperparameter combinations can be tried in parallel. When trying to leverage knowledge from multiple datasets, this approach has the nice property that it is possible to make sure that at least one of the default settings performs well on each benchmark dataset. Multiple defaults is a concept suggested by Wistuba et al. \cite{wistuba2015sequential}, later developed into a method based on greedy forward search by Pfirsterer et al. \cite{pfisterer2021learning}, who chose 6 algorithms from OpenML to test out their method on 100 datasets of between 500 and 100000 observations and up to 5000 features. Both papers did not use any neural network methods however, unlike our research which focuses specifically on a wide-kernel convolutional architecture. 

\subsubsection{Neural Architecture Search}

%Hyperparameters are extra decisions that accompany the an algorithm, making it possible to modify the algorithm before using it. 
%Hyperparameters  and understandably there are various types of hyperparameters. %and considering only the case of neural networks there is a distinction that can be drawn between architectural hyperparameters, deciding the shape and structure of the network, and training hyperparameters, deciding how the network is shown data and iteratively updated. 
Architectural hyperparameters make it possible to modify the shape and structure of a neural network before using it. 
Within AutoML, the niche that focuses on architectural hyperparameters is known as “neural architecture search” (NAS) \cite{baratchi2024automated} and it is often decomposed further into two types of decision: (1) choice of an overall structure and (2) choice of sub-modules. These are called macro and micro searches. %Macro searches are intended to combine layers (reusable modules) into larger neural networks. Micro searches instead focus more on designing new layers within a neural network. They
Micro searches can rely on a template for the full network, like ResNet which is highly repetitive, in order to take a module and convert it into a testable network. Avoiding the combinatorial problem of looking at all the ways different layers can combine, micro searches can potentially offer a smaller search space than macro searches are liable to. To pick just one example of many, NASNET is a micro search. This research considered options for ‘normal’ and ‘reduction’ components in convolutional networks in order to identify damage caused by COVID-19 in chest x-ray images, and ended up achieving 97\% accuracy \cite{martinez2020performance}. Our work focuses on architectural hyperparameters within a relatively fixed network architecture, reflecting adaptations of the network rather than the design of a completely new network. This allows us to keep within a restricted search space and focus on understanding hyperparameters, a crucial consideration that is generally overlooked in the neural architecture search literature. Another point to note is that our work focuses on time series data, which has been much less investigated than other types of data in the context of NAS \cite{martinez2020performance}. 

\subsubsection{Meta-Learning}

Amongst past work on hyperparameter optimisation, there is one more trend that is relevant to our current study. This trend is called “meta-learning” %and it is based on the idea of exploiting metadata in order to re-use information about how algorithms performed previous tasks. The 
and its goal is to predict which ML algorithm will perform a new task well, using a `meta' algorithm to make the decision. The key information used to make predictions is the observed performance of ML algorithms on past datasets along with “meta-features” which summarise (previous and new) datasets. Meta-features describe the properties of a dataset and can be used to say how similar two datasets are. %Meta-models predict how successful a given model will be when trained and evaluated on a given dataset. 
A trained ``meta-model'' can be used to either directly suggest an ML algorithm to try out on a new dataset, or it can be used to warm-start a search method such as Bayesian optimisation \cite{baratchi2024automated}. 

Different types of meta-feature exist, ranging from simple metadata like the number of variables measured, to parameters extracted from fitted statistical models, or even the performance scores of quick-to-evaluate models like linear regression \cite{rivolli2022meta}. One of the main goals of our research is to investigate how time series data properties relate to architectural hyperparameters. Probably the closest past research to our own is \cite{petelin2023towards}, which used meta-learning to predict the performance of algorithms in a time series prediction challenge. They extracted features specific to time series using the TSFresh and catch22 analysis packages. Similarly, in order to predict COVID-19 instances across different countries, \cite{talkhi2024using} also used time series features for meta-learning although that research focused more on simpler ML algorithms. 

A key distinction of our work is that we focus on two of the most understandable meta-features (sampling rate and presence of high frequencies) and we directly manipulate these within our datasets. Previous research into meta learning has relied on coincidental differences in existing datasets, without actively manipulating meta features like the sampling rate. Additionally, our research focuses on the domain of bearing fault detection in vibration data, within which the differences between datasets might be more subtle than they are in more generic meta-learning benchmarks, while at the same time presenting larger and more challenging datasets than most time series benchmarks for meta-learning.

\subsection{Summary of Own Previous Work}

In \cite{hoogen2023hyperparams} we investigated how choices of the hyperparameters for a specific deep learning architecture (wide-kernel CNN) affect its performance in classifying faults from vibration signals in industrial machines. This architecture consists of five convolutional layers including an initial `wide' layer with a large kernel size which then feed into a fully-connected classifier. See Section \ref{sec:method:widekernelarchitecture} for a detailed description. 
We were particularly interested in learning more about how performance changes across the hyperparameter space and so we  performed a large-scale analysis, training and testing many versions of the wide-kernel CNN with different configurations (e.g., number of filters, kernel size) on three different datasets. The research used a grid search to explore the hyperparameter space and then presented multiple pieces of follow-on analysis. 

Looking at how the distribution of test accuracy scores changes as a result of tweaking individual hyperparameters, we found that the number of filters in the later convolutional layers (3 to 5) has the biggest impact on performance, with an optimal value around 32. The kernel size in the first layer also appeared to be important for datasets with a high sampling rate, where larger kernels perform better. 
Another discovery was that high-performing hyperparameter settings could be unstable, i.e., small changes to hyperparameters can have a large impact on performance, highlighting the need for careful tuning.

%Overall, this study provided insights into how to configure a wide-kernel CNN for better fault detection in industrial machinery. We also identified interesting areas for future research, such as applying this approach to other signal classification tasks and different neural network architectures.

A second study increased the number of benchmark datasets by four and applied 12,960 variations of the wide-kernel CNN to seven different industrial vibration datasets \cite{hudson2024stay}. Once again, we analysed the grid search to identify how hyperparameter settings influence the network's performance and how to effectively tune them for new data.

Compared to the previous study~\cite{hudson2024stay}, a more thorough investigation using seven datasets confirmed that the most important hyperparameters for the network's performance are the number of filters in layers 3-5 and the kernel size in the first layer. New analysis focused on pairwise interactions between hyperparameters, where changing one hyperparameter might require adjusting another one for optimal performance. 
After looking at how hyperparameters influence one another, we suggested a specific order for tuning the hyperparameters to minimise the need to re-tune them later. This would be useful in situations where practitioners want to tune the hyperparameters one-at-a-time sequentially rather than using a full grid search. 
Tuning the hyperparameters in our suggested order leads to better performance compared to random or reverse order.

Compared to our past work, the current study is able to make a broader argument about the importance of hyperparameter tuning by considering two alternative architectures to a wide-kernel CNN. Namely, we now experiment with a long- short-term memory (LSTM) and a transformer architecture. This allows us to show that hyperparameters are important across numerous fault detection algorithms, and also that hyperparameter tuning is important when neural networks are used across seven different fault datasets. 

In addition, we investigate the impact of actively modifying dataset properties via resampling and filtering. This leads to new experiments with modified versions of one benchmark dataset. Ultimately our experiments with resampling and filtering lead to the new conclusion that sampling rate and the presence of high-frequency noise are not the main reason that a wide kernel is better than a short kernel, contrary to earlier speculation. 

Moreover, we also conclude that due to the impact of differences between datasets, there will be no single set of default hyperparameter settings that is always best. We therefore establish ``multiple defaults'' for our wide-kernel CNN architecture, an extremely efficient way to try out hyperparameters combinations on new data which we did not explore in our previous research.

\section{Method}\label{sec:method}

Our research method consists of several parts which feed into each other. In this section, we step through each piece of the method and describe it in detail. 
Starting out, we explain that we use seven different benchmark datasets in order to account for variations in the data used to train and test neural networks.
%Starting out, we describe seven different benchmark datasets that we use to train and test neural networks.
%Different types of fault, machinery and recording set-up are possible in fault detection and we therefore expect that datasets can differ from one another. We want to account for variations in the data and so we assembled a selection of seven benchmarks to train and test out CNNs. The properties of the benchmark datasets are described in detail below.  

Next, our method splits into two main structural components based on which architectures we study. Previous work in \cite{hudson2024stay} and \cite{hoogen2023hyperparams} looked at the consequences of tuning the hyperparameters of a wide-kernel CNN but not other kinds of architecture. We devote a short section to the impact of hyperparameter tuning on two other architectures with the goal of showing more generally that hyperparameters matter for bearing fault classification, at least when using neural networks. We introduce two popular types of architecture that are used in this short section as alternatives to a wide-kernel CNN.  
Additionally, we describe how we generate results by sampling different hyperparameter values from a range of possibilities and then training and testing networks on the seven benchmark datasets. 

The second structural component of the method and the majority of the research however focuses on the wide-kernel CNN and aims to investigate how much dataset properties affect which hyperparameter settings are best. We describe the wide-kernel CNN architecture used in our work and its architectural hyperparameters. 
Training and testing on seven benchmark datasets gives an indication of how differences in the data can have consequences for hyperparameter tuning. 
%We start off with a grid search which is an exhaustive exploration of different combinations of hyperparameters and for each combination we train and test a CNN on benchmark data. We repeat the grid search seven times in order to include all the benchmark datasets mentioned earlier. 

In addition to accepting the benchmark datasets as-is, we also want to get into a position to see what happens when we manipulate the dataset properties ourselves. By manipulating the datasets, we gain fine-grained control over factors that might explain why some hyperparameter settings result in high accuracy while others perform badly. In concrete terms, we edit the sampling frequency of the CWRU dataset by resampling it and we also cut out high frequencies by filtering. %Both types of data manipulation are described below. Each modified version of the CWRU data becomes the input to a new grid search over important hyperparameters and produces a new grid of test accuracy scores. Collating these grid searches, we can compare how hyperparameters behave in response to different levels of resampling or filtering. 

\begin{figure}[htb]
    \centering
    \includegraphics[width=1\columnwidth]{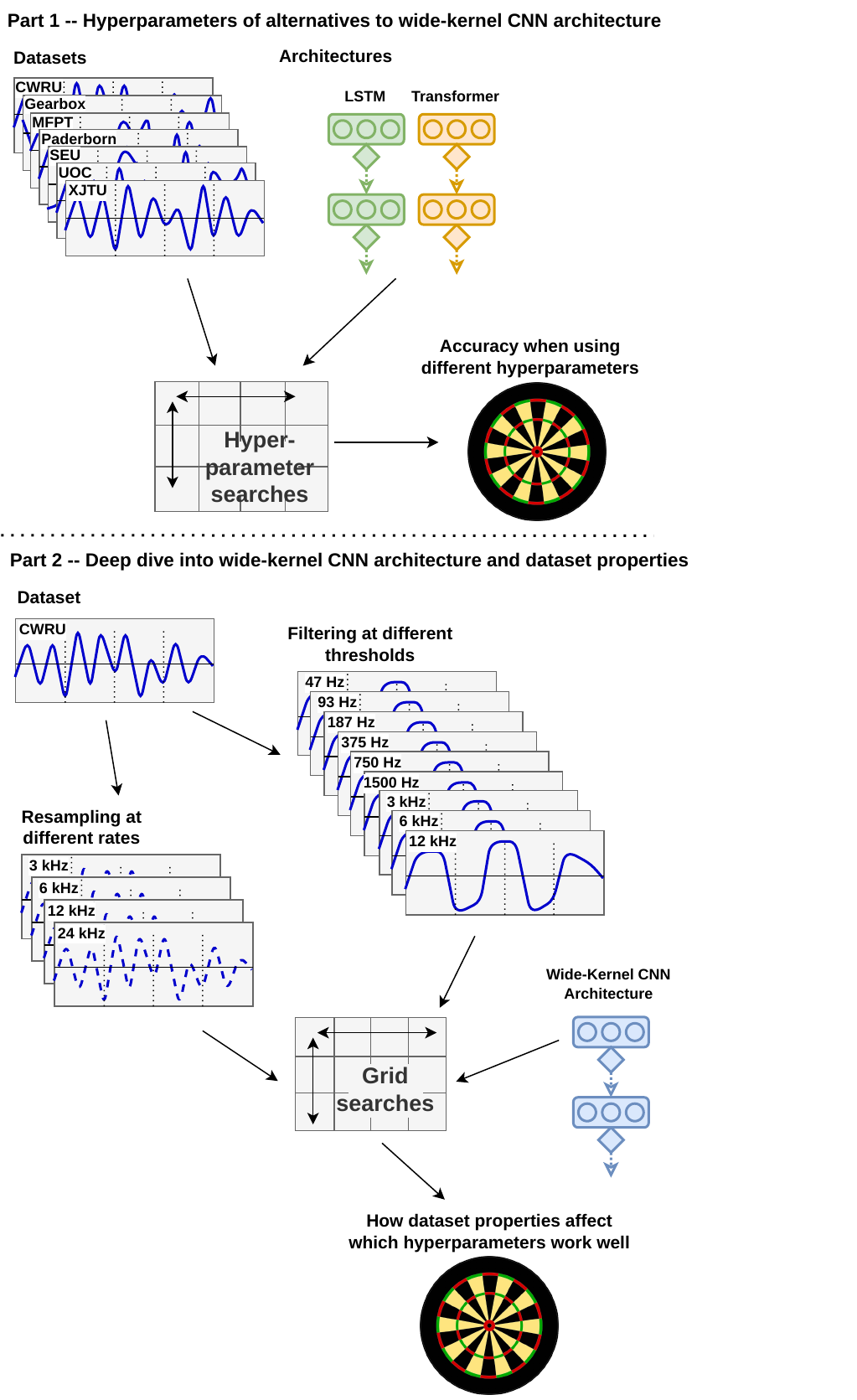}
    \caption{Workflow of the method to test out multiple architectures on varying data. 
    }
    \label{fig:workflow}
\end{figure}

At this point, we have several different types of architecture being tested out on different datasets; for clarity, we illustrate the overall flow of the various steps in Figure \ref{fig:workflow}.

%Ultimately we wish to extract useful insights into how to tune hyperparameters. Using the results from different benchmarks, we conduct multiple analyses. Firstly we show a descriptive analysis of the average performance and variability when changing hyperparameters of different architectures. Focusing on wide-kernel CNNs, we examine the interactions between hyperparameters. Then, we move on to identify multiple defaults to use on new data. A detailed analysis of the impact of data manipulation (resampling and filtering) rounds off the method. %There are actually several related analyses based on the outputs of the grid searches, and we go over each one in turn. Correlations are used to see how much hyperparameter settings generalise across datasets. 
%Feature importance is used to see which hyperparameters matter the most. 
% We have two approaches, the first of which is based on Shapley values and the second of which is based on the d-dimensional earth mover's score. What these terms mean is laid out towards the end of this section. 

\subsection{Datasets}

The most obvious input to our experiments is data. Different types of fault, machinery and recording set-up are possible in fault detection and we therefore expect that datasets can differ from one another. We want to account for variations in the data and so we assembled a selection of seven benchmarks to train and test out CNNs, the same seven that we used earlier in \cite{hudson2024stay}.
All of the datasets are benchmarks used for fault detection experiments in industrial machinery, focusing particularly on bearings and gearboxes. All of the datasets are described in more detail in Appendix \ref{sec:app:benchmarks}.

The CWRU bearing dataset derives from a structured experimental setup which mirrors real-world conditions with various fault locations and severity levels. There are minor imbalances in the class due to differing machine operating conditions. The Gearbox dataset is used to simulate real-time diagnostics, with balanced data for classifying healthy versus faulty gearboxes under varying load conditions.
The MFPT dataset explores faults in machinery with diverse fault types and unbalanced data due to different sampling frequencies. The XJTU datasets provide extensive run-to-failure data for bearing faults makes it suitable for analysing the remaining useful life.
The UoC dataset focuses on gear faults with a balanced setup, targeting multi-class classification tasks through single-channel vibrational data. Lastly, the SEU datasets offer insights into bearings and gearboxes with comprehensive multivariate data across various fault states and operational conditions.

Overall, these datasets are instrumental for developing and evaluating fault detection algorithms in machinery, each contributing unique features to test.

\subsection{Method Pt. 1 -- Experiments on Alternatives to Wide-Kernel CNN Architecture}

A wide-kernel CNN architecture takes centre stage in part 2 of our method, however we also want to say that hyperparameters are an important and unavoidable decision for other neural networks too. Therefore, in addition to the wide-kernel CNN which is our focus we also perform limited experiments on two additional architectures. By encompassing more than one architectural style, our work has a broader basis on which more general conclusions can be built. 

The first additional architecture uses ``long- short-term memory'' (LSTM) to digest the signal as a sequence of smaller pieces. The second additional architecture, a ``transformer'', also consumes its inputs as a sequence. Both architectures boast a good record of accomplishing sequence processing tasks and can be adapted to work on time series. 

%Our goals for the LSTM and transformer architectures are not the same as those we have for the wide-kernel CNN. 
%With the wide-kernel CNN, our intention is to weigh up the importance of different hyperparameters and familiarise ourselves with each hyperparameter in detail. Our results do not generalise to other architectures which have their own distinct sets of hyperparameters. The extra LSTM and transformer architectures are not included to allow for a detailed analysis 
In this way, the goal of part 1 of our method is to provide evidence that hyperparameter tuning has important consequences for a wide range of state-of-the-art neural networks. We want to show that the dangers of ill-hyperparameterisation are not a consequence of choosing the wide-kernel architecture in particular. 

\subsubsection{Long Short Term Memory (LSTM) Architecture}

Following the example of \cite{chen2021bearing}, we consider a bearing fault classifier built around Long Short Term Memory (LSTM) layers. The central idea of an LSTM layer is that it passes information forward between steps when processing a sequence stepwise -- making it possible to `remember' things seen earlier -- and yet it also allows the internal flow of information to be flushed out \cite{hochreiter1997long}. Strategically flushing out the internal information deals with the problems of vanishing and exploding gradients that previous kinds of recurrent network suffered from, with the upshot being that LSTMs are more successful at handling longer sequences and performing a wide range of tasks.

Like \cite{chen2021bearing}, we place an LSTM in an overall architecture that has two CNN components feeding into the LSTM itself. The two CNNs are used in order to achieve a ``multi-scale'' effect, with one CNN containing smaller kernels and the other containing larger kernels. Both 
CNNs summarise the raw inputs into a more concise sequence, which is needed because the LSTM layer would be quickly overwhelmed otherwise. The LSTM component is a pair of repeated layers. After the LSTM, fully connected layers lead to a final prediction of what type of fault is present in the machinery. The architecture is visualised in Appendix \ref{sec:app:lstm}.

\subsubsection{Transformer Architecture}

We also consider a transformer architecture, following the example set by \cite{zhao2023fault} for bearing fault analysis. 
Transformers use a so-called `attention' mechanism which allows them to process each step in a sequence by pooling information from other steps \cite{vaswani2017attention}. The innovation of using attention provides an efficient alternative to recurrent connections, and the booming popularity of transformers in applications like ChatGPT testifies to their efficacy.

The network we use receives incoming signals in a CNN component which leads to a compact sequence that then passes through to the transformer component. 
The CNN has only two layers and it shrinks the raw signal to a sequence of roughly 1/16th the length. 
The transformer section of the architecture consists of 4 layers by default. 
Global average pooling layer is used to consolidate the outputs from the transformer into a single vector. Finally there is a single fully-connected layer which outputs the final prediction of the network. The architecture is visualised in Appendix \ref{sec:app:transformer}.

\subsubsection{Hyperparameter Search on LSTM and Transformer Architectures}

When studying the LSTM and transformer networks we restrict ourselves to drawing some generic conclusions about the impact of (mis)tuning hyperparameters. As such, we do not find it necessary to perform a full grid search, but instead to rely on a reduced sample from a grid of possibilities. We select 100 configurations uniformly at random from a larger collection of possible hyperparameter combinations. 

For the LSTM-based neural network, the values we consider for each hyperparameter are shown in Table \ref{tab:hyperparametersSummaryLSTM}. The middle value for each hyperparameter is based on what it is set to by default in the code published alongside \cite{chen2021bearing}. The other two options simply expand or shrink the amount by a factor of four. From the space of possibilities defined in Table \ref{tab:hyperparametersSummaryLSTM}, we take 100 random samples. 

\begin{table}[htb] 
\centering
\footnotesize
\caption{Hyperparameter values considered for the LSTM network.\label{tab:hyperparametersSummaryLSTM}}

\begin{tabular}{@{}ll@{}}
\toprule
 \textbf{Hyperparameter}  &\textbf{Value Domain} \\
 
\midrule
Kernel size shallow network layer 1  & \{5, 20, 80\}  \\
\midrule
Filters shallow network layer 1  & \{13, 50, 200\}  \\
\midrule
Kernel size shallow network layer 2  & Always half of layer 1 value  \\
\midrule
Filters shallow network layer 2  & \{8, 30, 120\}  \\
\midrule
Kernel size deep network layer 1 & \{2, 6, 24\}  \\
\midrule
Filters deep network layer 1  & \{13, 50, 200\}  \\
\midrule
Kernel size deep network layer 2 & Always equal to layer 1 value \\
\midrule
Filters deep network layer 2  & \{10, 40, 160\}  \\
\midrule
Kernel size deep network layer 3 & Always equal to layer 1 value  \\
\midrule
Filters deep network layer 3 & \{8, 30, 120\}  \\
\midrule
Kernel size deep network layer 4 & Always equal to layer 1 value  \\
\midrule
Filters deep network layer 4 & Always equal to filters in layer 2 \\ & of the shallow network  \\
\midrule
Hidden units LSTM layer 1 & \{15, 60, 240\}  \\
\midrule
Hidden units LSTM layer 2 & \{15, 60, 240\}  \\

\bottomrule
\end{tabular}
\end{table}

The transformer hyperparameters are also confined to a limited set of possibilities, which are shown in Table \ref{tab:hyperparametersSummaryTransformer}. From these we choose a sample of 100 combinations of values. Like with the LSTM, our choice of default options is inspired by the code published by the authors (alongside \cite{zhao2023fault}).

\begin{table}[htb] 
\centering
\footnotesize
\caption{Hyperparameter values considered for the transformer network.\label{tab:hyperparametersSummaryTransformer}}

\begin{tabular}{@{}ll@{}}
\toprule
 \textbf{Hyperparameter}  &\textbf{Value Domain} \\
 
\midrule
Kernel size CNN layer 1  & \{4, 16, 64\}  \\
\midrule
Filters CNN layer 1  & \{1, 4, 16\}  \\
\midrule
Kernel size CNN layer 2  & \{4, 16, 64\} \\
\midrule
Filters CNN layer 2  & \{2, 8, 32\}  \\
\midrule
Hidden units in transformer layers & \{64, 256, 1024\}  \\
\midrule
Attention heads in transformer layers & \{8, 32, 128\}  \\
\midrule
Number of transformer layers & \{1, 4, 16\}  \\

\bottomrule
\end{tabular}
\end{table}

\subsubsection{Analysis Techniques}

Our analysis of the LSTM and transformer architectures asks how much the accuracy of a network changes when its hyperparameters change. The evidence considered is a sample of 100 LSTMs and 100 transformers chosen at random from the range of hyperparameter options, with each network trained and evaluated on seven benchmark datasets. Based on their hyperparameters, some of the 200 networks perform better and some perform worse. 
When we report the results later, we encapsulate the distribution of accuracy scores (per benchmark) through Tukey's five-number summary. 
%To give an overview of the results, we present to the reader a summary of the distribution of accuracy scores. We report the distribution of test accuracy -- the spread of how well the networks performed -- as follows. First, for the LSTM we state the minimum, 25th percentile, median, 75th percentile and maximum accuracy achieved after training and testing on CWRU data. We then state what values the LSTM reached on the Gearbox data for the minimum, 25th percentile, median, 75th percentile and maximum (also known as Tukey's five-number summary). The results proceed to give the corresponding values for all the other benchmark datasets and when using the transformer instead. In total, there are 5 values describing how 2 architectures did on 7 different datasets, with the goal being to give an idea of the spread of the data and quickly clarifies how much the architectures are affected by their hyperparameters and by changes in the data. 
We are also curious how well the default hyperparameter values used in past literature fare. Therefore we also report the accuracy obtained when using the default settings.

\subsection{Method Pt. 2 -- Focusing on the Wide-Kernel CNN Architecture}

Next we leave behind the LSTM and transformer architecture and arrive at the main body of the research, which is to explain how to set the hyperparameters of a wide-kernel CNN architecture. 

\subsubsection{Overview of  Architecture}\label{sec:method:widekernelarchitecture}

Previous work demonstrated excellent performance in bearing fault detection by employing a wide-kernel in the first convolutional layer followed by smaller kernels in subsequent layers~\cite{hoogen2020improvedWDCNN,zhang2017new,hoogen2021,hoogen2023hyperparams}. This wide-kernel CNN does not only perform well but is also particularly suitable for fault detection tasks due to its reasonable size in terms of convolutional layers and its capability to process raw time series directly.

Our architecture contains two convolutional layers, each utilising several combinations of stride, kernel size and filters. 
After that, a \textit{convolutional unit} (see Figure~\ref{fig:architecture}) is repeated \textit{three} times, resulting in a total network of five convolutional layers, as described in previous work~\cite{hoogen2020improvedWDCNN,hoogen2021,zhang2017new}. 
After each convolutional layer, the network utilises local average pooling to decrease the vector size of the convolutional output. 
At last, the reduced output from the convolutional layers is fed into fully connected layers that classify the vibrations. See Figure~\ref{fig:architecture} in Appendix \ref{sec:app:architectures} for a visual depiction of the architecture, and additional supplementary details. 

We originally chose to use a wide-kernel CNN architecture as the focus of our hyperparameter investigation because it had already been achieved near-perfect accuracy on the two bearing fault detection datasets (CWRU and Paderborn) in prior work~\cite{zhang2017new,hoogen2020improvedWDCNN,hoogen2021}. Its performance on these benchmarks indicated to us that it is well-suited for fault detection tasks and that a more complicated network would not necessarily give extra benefit. Furthermore, the architecture is lightweight in terms of computational resources and memory usage, which allows it to have a wider range of uses, in particular opening up the possibility of edge computing in the context of Industry 4.0. 

Moreover, the modest depth of the wide-kernel architecture, compared to other state-of-the-art CNN architectures~\cite{periodnet, zhao2023fault,qiang2023intelligent,CHEN202054, chen2021bearing, zhou2024faultformer}, lends itself well for an extensive grid search of the hyperparameters due to the relatively low number of hyperparameters.

\subsubsection{Hyperparameter Search}

Each hyperparameter of a neural network is a decision and changing any one of them leads to a new configuration for the network, potentially leading to a different accuracy score on the task being attempted. From any given starting point there are as many directions for the configuration of the network to go as there are hyperparameters. By imposing lower and upper limits on what values each hyperparameter can take, it is possible to chart out a finite hyperparameter space for exploration. In the ensuing section of this paper, we describe the hyperparameter spaces used in our different experiments and how we explore them. 

Earlier work on the wide-kernel CNN identified seven architectural hyperparameters \cite{hoogen2023hyperparams}. These were explored using a grid search strategy in which each hyperparameter is restricted to a limited set of values and every possible way of combining these is tested. For example, every kernel size in the first layer is tried out with every stride in layer 1, and each of these combinations is tried out with every kernel size in layer 2, etc. 
Since they differ from one another, the set of options tried for any particular hyperparameter is not the same as for other hyperparameters. Table \ref{tab:hyperparametersSummary} shows the options. Looking at all the mixtures that can be created, there are $5 \times 3 \times 6 \times 2 \times 6\times 2\times 6 = 12960$ combinations included in the grid search.

Previous work also found that three hyperparameters in particular are important: kernel size in the first layer, number of filters in the first layer and number of filters in layers 3-5 (these are indicated in italics in Table \ref{tab:hyperparametersSummary}) \cite{hoogen2023hyperparams,hudson2024stay}. 
We use these hyperparameters as a basis for new grid searches.
The reason for new grid searches is that we want to find out more about the impact of manipulating the properties of the data by filtering or resampling. 
In the next subsections we describe our approach to filtering and resampling the data to create new versions of it %is described in the next subsections. 
%By manipulating the dataset we create new versions of it. 
On each version, we perform a grid search considering the values shown in Table \ref{tab:hyperparametersSummary}. %The data manipulation strategies and the goals of performing them are explained in more detail in the next subsection. 

In summary, we have a grid search performed over seven hyperparameters which tests the accuracy of the network on seven benchmark datasets. Additionally, as will be explained in a moment, we have multiple smaller grid searches which apply to modified versions of the data. These smaller searches operate over three hyperparameters. 

\begin{table}[htb] 
\centering
\footnotesize
\caption{Architectural hyperparameters with the values they take in grid search.\label{tab:hyperparametersSummary}}

\begin{tabular}{@{}ll@{}}
\toprule
 \textbf{Hyperparameter}  &\textbf{Value Domain} \\
\midrule
\textit{Kernel size layer 1  }& \{16,  32,  64, 128, 256\}  \\
\midrule
Stride layer 1  & \{4,  8, 16\}  \\
\midrule
\textit{Filters layer 1}   & \{8,  16,  32,  64, 128, 256\} \\
\midrule
Kernel size layer 2  & \{3, 6\} \\
\midrule
Filters layer 2  & \{8, 16, 32, 64, 128, 256\}  \\
\midrule
Kernel size layers 3-5  & \{3, 6\}  \\
\midrule
\textit{Filters layers 3-5 } & \{8, 16, 32, 64, 128, 256\} \\
\bottomrule
\end{tabular}
\end{table}

\subsubsection{Analysis Techniques on Seven Benchmark Datasets}

In order to flesh out certain points from the work reported earlier in \cite{hoogen2023hyperparams, hudson2024stay}, we take the results from a grid search over seven hyperparameters applied to seven benchmark datasets and present them in new ways, as follows. 

\vspace{1 em}
\noindent
\textit{Descriptive Analysis}: 

\noindent If possible, we would like to recommend effective default values for the seven architectural hyperparameters of our wide-kernel CNN architecture. Recommendations would be easy to make if we find that a single value performs best across all the seven benchmark datasets. Therefore, we present a simple table that shows the average accuracy obtained on each dataset when using each hyperparameter value. The table compares different options for a hyperparameter, showing if there is a generally best option or if instead there is no %obvious default v
value that performs well on all datasets. 

\vspace{1 em}
\noindent
\textit{Influence between Important Hyperparameters}: 

\noindent Part of the work of \cite{hudson2024stay} was to attribute importance scores to the hyperparameters after testing out hyperparameter configurations on seven benchmark datasets. These importance scores represented the extent to which a hyperparameter impacts the accuracy of the wide-kernel CNN. In particular, the results identified three especially important hyperparameters: the number of filters in the first layer, the kernel size in the first layer, and the number of filters in layers 3-5. 

In the aforementioned past work there was also an analysis of the influence each hyperparameter on other hyperparameters. The `influence' of A on B was interpreted as the likelihood that the optimal value for B will change as a result of tuning A. In other words, A is influential on B if tuning A is likely to mean that B needs re-tuning. 

When someone tunes a network, they tweak the options available to them until they find hyperparameter settings that result in better accuracy. One approach to the problem is to focus on a single choice -- for example the number of convolutional filters used in the first layer of the network -- and modify this value without changing anything else. We call this approach tuning a hyperparameter `individually'. 
Tuning individually reduces the number of possibilities, perhaps to testing out a handful of values the hyperparameter might reasonably take.  
After trying the possibilities the best performer will be retained. 

An underlying assumption of individually tuning hyperparameters is that there is a definite starting point. All of the other hyperparameters must be given a fixed value at the beginning of the process. 
The results from individual tuning are therefore contingent on there being a starting point and indeed we may find that a different starting point leads to a different optimal value for the hyperparameter being tuned. 
Even if a hyperparameter has been given the `best' value through individually tuning it, we may find that the best value changes when another hyperparameter has been edited. In which case, the original hyperparameter needs to be re-tuned to reflect a new starting point. 

It is not clear whether changing one hyperparameter will have any impact on another. It is possible for hyperparameters to be largely independent so that we can modify one of them freely without it affecting which value is optimal for the other hyperparameter. Alternatively, they could be closely linked and it might be best to tune them so they always change together. In our research we consider the derived question of whether tuning one hyperparameter individually has an impact on the outcome from tuning another hyperparameter. The answer will be a probability because we need to summarise what happens across a large range of starting points. We have seven hyperparameters and each way of setting them is a different starting point. 

We therefore revisit the notion of `influence' used in \cite{hudson2024stay} and computed by Algorithm \ref{alg:influence}. This fairly straightforward method to quantify influence considers every combination of values in a grid search as a starting point. From each starting point the algorithm tests the influence of individually tuning one hyperparameter (let us call it `A') on another hyperparameter (`B'). First, B is tuned, then A is tuned and subsequently B is re-tuned to see if the optimal value has changed. If re-tuning changes the value of B then the algorithm increments a running count. At the end, the algorithm returns the final value of the running count divided by the number of starting points that were tested. 

\begin{algorithm}
\caption{Compute the influence of $A$ on $B$}\label{alg:influence}
\begin{algorithmic}
\STATE trialCount $\gets$ 0
\STATE differenceCount $\gets$ 0
\FOR{\textbf{each} configuration $\mathbf{c} \in (\mathbf{c_1}, \mathbf{c_2}, ..., \mathbf{c_{12960}})$}
    \STATE $B_{tuned} \gets$ tune($\mathbf{c}$, $B$)
    \STATE $\mathbf{c'} \gets \mathbf{c}$
    \STATE $\mathbf{c'}[A] \gets$ tune($\mathbf{c'}$, $A$)
    \STATE $B_{re-tuned} \gets$ tune($\mathbf{c'}$, $B$)
    \STATE trialCount $\gets$ trialCount $+ 1$
    \IF{$B_{tuned} \neq B_{re-tuned}$}
        \STATE differenceCount $\gets$ differenceCount $+ 1$
    \ENDIF
\ENDFOR \textbf{ each}
\RETURN differenceCount $/$ trialCount 
\end{algorithmic}
\end{algorithm}

When reporting these influence scores, \cite{hudson2024stay} provided a visualisation to show the influence each hyperparameter has on one another. 
In the current paper, we repeat this procedure but consider what happens when only the three most important hyperparameters are tuned and the other hyperparameters are held to their default values (as defined in \cite{hoogen2020improved}). 
We show this new information since it allows us to focus on strategies that allow for faster and more effective tuning. Excluding hyperparameters that have low impact on the accuracy of the CNN makes it possible to cut down on computation. 

\subsubsection{Multiple Defaults}

Similar to previous work \cite{pfisterer2021learning}, we pursue the idea of ``multiple defaults''. Rather than providing a single combination of hyperparameters we aim to provide a series of hyperparameter combinations to try out. Our goal is that, after trying $n$ elements of the series, we have as high a chance as possible that at least 1 of the hyperparameter combinations performs well on our data. Different defaults might perform well on different datasets. We now define the task of finding multiple defaults.

For algorithm $A$, let $\Theta$ be a hyperparameter space defined by hyperparameters $\Theta_1, \Theta_2, \ldots, \Theta_n$, and $\theta$ denote a `combination' that specifies a value for each hyperparameter. Each possible $\theta$ provides all the hyperparameters needed to run the algorithm and can be thought of as a point within $\Theta$. 

We intend to find multiple $\theta$ which will be the multiple defaults. To distinguish them, we index them $\theta_1, \theta_2, \ldots$ where each $\theta$ is a different combination of hyperparameter settings, i.e., a different point in the hyperparameter space. Each combination is itself composed of multiple values, one value for each hyperparameter. To distinguish the values that compose a single combination, we introduce the following notation. Let hyperparameters within combination $\theta_x$ be written as $\theta_{x,1}, \theta_{x,2}, \ldots, \theta_{x,n}$. For the $x$th hyperparameter combination, the value of the $y$th hyperparameter would be $\theta_{x,y}$. 

Our search for multiple defaults will be guided by how well the hyperparameter combinations perform on different benchmark datasets. Let us say there are $l$ different benchmarks. In our research there are seven benchmark datasets, i.e., $l=7$. Each benchmark acts as a function on hyperparameter combinations, written below as: $b_1(\theta), b_2(\theta), \ldots, b_l(\theta)$. Each function takes a hyperparameter combination as an input and returns a real number which is the percentile of how well that combination performed compared to other combinations. 

Next we describe the search for multiple defaults. The search proceeds iteratively through multiple steps. At each step of the search, the goal is to identify which hyperparameter combination should be added to the series in order to increase (as much as possible) the maximum performance of any hyperparameter combination seen so far. The maximum performance of hyperparameter combinations $\theta_1, \theta_2, \ldots, \theta_m$ on benchmark 1 is $max(b_1(\theta_1), b_1(\theta_2), \ldots, b_1(\theta_m))$. At the end of step $m$, we have chosen $m$ hyperparameter combinations. With combinations $1, \ldots, m$ and benchmarks $1, \ldots, l$, the expected best performance of the $m$ combinations is

\begin{equation} \label{eq1}
\footnotesize
\begin{split}
E_m = \mathbb{E}\Big| \max \Bigl( b_1(\theta_1), \ldots, b_1(\theta_m) \Bigr), \max \Bigl( b_2(\theta_1), \ldots, b_2(\theta_m) \Bigr), \ldots, \\\max \Bigl( b_l(\theta_1), \ldots, b_l(\theta_m) \Bigr) \Big|
\end{split}
\end{equation}

\normalsize
The choice of the $m$th hyperparameter combination is the one which increases the expected best performance of the previous $m-1$ combinations the most. 

Previously, a greedy algorithm was suggested for finding multiple defaults \cite{pfisterer2021learning}. The greedy approach is relatively efficient to compute but it does not guarantee the global optimum. By contrast, since we have a completed grid search to work with, we are able to take an exhaustive approach and thus find the globally best-performing series. Because the completed grid search already contains the performance scores (for each combination of hyperparameters on each benchmark) it is possible to execute a brute force algorithm. The computationally expensive step of training and evaluating the CNNs has already been done, making the brute force search a nested loop over the table of results. 

In theory, we could keep on adding more and more hyperparameter combinations into our series of multiple defaults. However, in practice we find that the search will identify a relatively small series of hyperparameter combinations that cannot be improved further. Therefore, we terminate our search after $m+1$ steps when we find that $E_{m+1} = E_m$, and we return the series found at the end of the first $m$ steps (we return $\theta_1, \theta_2, \ldots, \theta_m$).

There is one final follow-on analysis which we include in order to evaluate how well the method generalises to new datasets. In this extra analysis we perform multiple runs of the search using leave-one-out validation. When there are $l$ benchmarks, $l-1$ benchmarks are used to determine defaults and then the last benchmark is employed to evaluate their effectiveness. This cycle is repeated $l$ times, ensuring each benchmark is omitted once. Within each repetition we calculate the best performance when applying defaults to the unseen benchmark. Averaging across the $l$ repetitions of the cycle gives an estimated performance on new data. 

In summary, our search for multiple defaults leads to $m$ different combinations of hyperparameters, arranged in a sequence. On new data, these combinations can be tried out in turn. 
If we do not try out all $m$ combinations but only $k$ of them, we still want to maximise the chance of finding a combination that works well. In our research we use seven benchmark datasets to evaluate how well different hyperparameter settings perform. After trying $k$ combinations, for each benchmark we see what did best. The best score per benchmark can be averaged, giving an idea of how quickly the procedure finds at least one successful hyperparameter combination on new data. 

\subsubsection{Resampling}

When tuning the first layer of the wide-kernel CNN we often find that the largest kernel size is best, but sometimes a small kernel is best and sometimes it does not matter. It depends what benchmark dataset is being used and therefore seems to be a result of differences in the data. We try to isolate which dataset properties cause different kernel sizes in the first layer to be better or worse. This leads to two types of data manipulation, the first of which is resampling. 

One practical question about how to tune the hyperparameters of a wide-kernel CNN is whether to resize the convolutional filters, particularly in the first layer, when the sampling rate changes. 
Fortunately, the sampling rate is something that we can control artificially. We resample windows of vibration data to a lower rate, for example going from an initial window sampled at 48 kHz to a window of data which now contains the same content but is sampled at 24 kHz instead. The highest frequencies will be lost, but otherwise the signal will be preserved with high accuracy. Simply it will take fewer data points to record the same back-and-forth vibrations, and each data point will represent a bigger jump forward in time from the previous one. 

To investigate the impact of the sampling rate on hyperparameters such as the kernel size of the first layer, we make multiple copies of the CWRU data and resample them by a factor of 2, 4, 8 and 16. Starting off with data sampled at 48 kHz, this results in versions sampled at 24 kHz, 12 kHz, 6 kHz and 3 kHz. As resampling increases, the inputs to the neural network get smaller. So we can resample more times before the inputs are too small for our CNN to handle, we start with windows of size 4096 for the resampling experiments, rather than 2048 (the default window size, used elsewhere in this paper, and described in Appendix \ref{sec:app:benchmarks}). A larger window size results in fewer examples to train from, which makes training more difficult and leads to lower average accuracy. 
On each version of the data, we run a grid search over the hyperparameters which previous work found most important. The grid search results can be analysed to compare what hyperparameter settings work best at different sampling rates.

For each combination of possible hyperparameters in our grid search, we train and test on the original and the four resampled versions of the CWRU 48 kHz dataset. The CWRU data was chosen because it combines: a relatively high sampling rate, a well-documented and high-quality data collection procedure, remaining a moderate challenge for neural networks, and being perhaps the most widely used benchmark in the field.

\subsubsection{Filtering}

It has been suggested in past work that the reason a wide kernel boosts the accuracy of a neural network is that wide kernels are good at filtering out high-frequency noise. 
We investigate this by employing a lowpass filter to deliberately remove high frequencies from the data. If the main accomplishment of a wide kernel is to filter out high frequencies, then a larger range of kernel sizes should become viable if the high frequencies are already eliminated by pre-filtering. 
We do not know what (if any) cutoff should be used to decide which frequencies to keep and which to throw away, so we try multiple cutoff thresholds, starting from near the upper limit provided by the Nyquist frequency. 
In real terms, given data like CWRU sampled at 48 kHz, we apply filtering to multiple copies of the data. We filter with cutoff frequencies of 12 kHz, 6 kHz, 3 kHz, 1500 Hz, 750 Hz, 375 Hz, 187 Hz, 93 Hz and 46 Hz. 
An additional benefit of using many cutoff thresholds is that we can also check if any of them are too drastic, shedding light on which frequencies are essential for accurate fault detection and cannot be removed. 

It should be noted that filtering in this way does not impact the sampling rate or the size of a window of vibration data, it simply removes content which occurs at high frequencies. This makes it easy to compare the results when performing a grid search over the most important hyperparameters. A full grid search over hyperparameters can be performed on each filtered version of the CWRU dataset, and the results are directly comparable since the same architectures are applied to data which has exactly the same size and shape.

\subsubsection{Analysis Techniques on Resampled and Filtered Data}

We analyse the resulting information in multiple ways. First, we look at correlations between different versions of the dataset. %, which gives an indication of the impact that resampling had. If resampling had no impact at all, then we would expect to have a perfect correlation of 1.0, and values lower than this would imply that resampling has an influence on which hyperparameter combinations are best. % stronger and which are weaker. 
Afterwards, we look at how well different kernel widths in the first layer perform. %This lets us question whether a wide kernel is desirable because of its ability to handle high-frequency noise, or whether another explanation is needed. %how important different hyperparameters are. We do this by initially asking a related question of whether the test accuracy of a network can be predicted from the hyperparameters it has. If the accuracy can be predicted, we can hone in on the specifics of how much each hyperparameter contributes to the predictability. 

\vspace{1 em}
\noindent
\textit{Correlation analysis}: 

\noindent By resampling the CWRU data, we create multiple versions of the dataset. One obvious question that arises is whether a combination of hyperparameters that works well on one version of the data will also work well on another. A similar question could be asked about the worst-performing hyperparameters. In general, this is a question of correlation: to what extent do the accuracy scores obtained on one dataset version correlate to the scores on data resampled to a different rate? If resampling had no impact at all, then we would expect to have a perfect correlation of 1.0, and values lower than this would imply that resampling has an influence on which hyperparameter combinations are best. We compute and then display the correlations in a correlation plot, which gives an insight into how much the hyperparameters are affected by the resampling/filtering. 
%One possibility is that certain hyperparameter combinations are simply better than others, and it does not matter if the properties of the data (such as sampling rate) change. In the extreme, this would result in a correlation of 1.0 between all versions of the dataset. However, if smaller correlations are found, then it would suggest that the properties of the data impact which hyperparameters are best, or that there is some degree of randomness in the accuracy scores that deflates the correlations. 

\vspace{1 em}
\noindent
\textit{Box plots}: 

\noindent One of the key questions in our research is whether the optimal kernel size in the first layer changes when the data is either resampled or filtered. Box plots provide a straightforward view onto how well different kernel sizes tend to work. We plot the distribution of accuracy scores for each kernel size when applied to each version of the data. This lets us question whether a wide kernel is desirable because of its ability to handle high-frequency noise, or whether another explanation is needed.

\subsection{Summary of Datasets Used}

Across the various parts of our research, multiple datasets are employed. For the sake of clarity we now present the full collection of datasets we use and remark on which architectures they are used with. Table \ref{tab:datasets_summary} shows these details.

\begin{table}[ht!]
\noindent
\scriptsize
\centering
\begin{tabular}{lr}
\toprule
\textbf{Dataset} & \textbf{Used With} \\
\midrule
\textit{Unmanipulated benchmarks} \\
\midrule
CWRU      &    LSTM, Transformer, Wide-kernel CNN \\
Gearbox   &    LSTM, Transformer, Wide-kernel CNN \\
MFPT      &    LSTM, Transformer, Wide-kernel CNN \\
Paderborn &   LSTM, Transformer, Wide-kernel CNN \\
SEU       &    LSTM, Transformer, Wide-kernel CNN \\
UOC       &    LSTM, Transformer, Wide-kernel CNN \\
XJTU      &     LSTM, Transformer, Wide-kernel CNN \\
\midrule
\textit{Resampling }\\
\midrule
CWRU 48 kHz resampled to 24 kHz & Wide-kernel CNN \\
CWRU 48 kHz resampled to 12 kHz & Wide-kernel CNN \\
CWRU 48 kHz resampled to 6 kHz & Wide-kernel CNN \\
CWRU 48 kHz resampled to 3 kHz & Wide-kernel CNN \\
\midrule
\textit{Filtering} \\
\midrule
CWRU 48 kHz lowpass at 12 kHz & Wide-kernel CNN \\
CWRU 48 kHz lowpass at 6 kHz & Wide-kernel CNN \\
CWRU 48 kHz lowpass at 3 kHz & Wide-kernel CNN \\
CWRU 48 kHz lowpass at 1500 Hz & Wide-kernel CNN \\
CWRU 48 kHz lowpass at 750 kHz & Wide-kernel CNN \\
CWRU 48 kHz lowpass at 375 kHz & Wide-kernel CNN \\
CWRU 48 kHz lowpass at 187 kHz & Wide-kernel CNN \\
CWRU 48 kHz lowpass at 93 kHz & Wide-kernel CNN \\
CWRU 48 kHz lowpass at 46 kHz & Wide-kernel CNN \\

\bottomrule
\end{tabular}
\caption{Summary of datasets used in the research.}
\label{tab:datasets_summary}
\end{table}

\section{Results}

Here we analyse the outputs from our grid search over hyperparameters, which has been applied to multiple datasets and to multiple filtered and resampled versions of the CWRU dataset. Combining these analyses leads to a broader understanding of how to set hyperparameters under different conditions. Before doing that however, we first attempt to establish the importance of hyperparameter tuning outside the context of a wide-kernel CNN by looking at how alternative architectures perform while their hyperparameters change. 

\subsection{Results Pt. 1 -- Experiments on Multiple Architectures}

Before we focus on a specific wide-kernel CNN architecture in part 2 of our results, we first 
%The overall focus of our work is a specific wide-kernel CNN architecture. However, %it is important to note that 
%other architectures also present the user with hyperparameter decisions, and also suffer from the fact that poor hyperparameter choices can be disastrous for accuracy. We therefore briefly 
present some experiments with two other popular architectural styles: a recurrent LSTM type of neural network and a transformer network. These results establish the general importance of hyperparameter tuning when working with neural networks.%, which is the basis for the remaining sections where we deal with the wide-kernel architecture much more extensively. 

\subsubsection{Long Short Term Memory (LSTM)}
Across our sample of 100 LSTM configurations, 
we see that poor hyperparameters can be disastrous. In the worst cases (CWRU and XJTU) the lowest-performing network was wrong more than 10 times as often as the best network. 
On all datasets the default values do roughly as well as about 75\% of modified hyperparameter settings and therefore are reasonable choices. 
One interesting thing is that the LSTM network’s default values appear to under-perform on XJTU data. Facing this benchmark, an LSTM configured with the defaults only achieved an accuracy of 72\% whilst it was possible to get above 90\% with other hyperparameter values. Re-tuning the default hyperparameter settings from the original authors seems to be a good idea on XJTU data, indicating that hyperparameters that perform well on one dataset do not necessarily perform well on another dataset. The default values fared relatively much better on CWRU and SEU. Overall the results show that hyperparameter tuning has a noticeable impact on the LSTM architecture much like with the transformer and wide-kernel CNN networks. 

\begin{table}[htb]
\noindent
\centering
\footnotesize
\begin{tabular}{lrrrrrr}
\toprule
{} &   Min &   25\% &  Median &   75\% &   Max &  Default \\
\midrule
CWRU      &   35 &   59 &      68 &   80 &   94 &           81 \\
Gearbox   &    6 &    9 &      11 &   13 &   17 &           14 \\
MFPT      &   46 &   49 &      50 &   50 &   54 &           52 \\
Paderborn &   28 &   40 &      46 &   50 &   57 &           48 \\
SEU       &   62 &   88 &      96 &   98 &   99 &           94 \\
UOC       &   24 &   35 &      41 &   50 &   72 &           54 \\
XJTU      &   26 &   48 &      57 &   75 &   97 &           72 \\
\bottomrule
\end{tabular}
\caption{Tukey's five-number summary of how accurate the LSTM architecture was with different hyperparameter settings. The accuracy score when using the default hyperparameter values is also shown in the right-hand column.}
\label{tab:lstm_accuracies}
\end{table}

\subsubsection{Transformer}
The transformer architecture performs relatively well when using the default hyperparameter settings, as shown in Table \ref{tab:faultformer_accuracies}. On four of the baselines it was able to correctly identify fault conditions on more than 90\% of test cases. Looking across all of the baseline datasets, the default hyperparameter settings were at least as good as 75\% of altered versions of the settings. The results suggest that the default values are quite reasonable choices. On the Gearbox dataset the transformer was inaccurate no matter if the default hyperparameter settings were used or not, suggesting a specific problem with that one dataset. 
Importantly, our results show overall that the transformer architecture is sensitive to hyperparameter tweaking. On CWRU and UOC for example, the accuracy plunges by roughly 90\% when swapping the best-performing hyperparameters for the worst.

Overall, the transformer architecture beats the LSTM architecture so long as it is given the right hyperparameter settings. The transformer was more sensitive to changes in the hyperparameters our study explored, in the sense that it could be very accurate or be very inaccurate depending on whether those were right. However, getting the hyperparameters right paid off for the transformer since its max accuracy was 40\% higher than the LSTM’s max when testing on UOC data. Given the correct hyperparameter settings, the transformer could be at least as good as any LSTM network we tried out on the seven benchmarks.

\begin{table}[htb]
\noindent
\footnotesize
\centering
\begin{tabular}{lrrrrrr}
\toprule
{} &   Min &   25\% &  Median &   75\% &   Max &  Default \\
\midrule
CWRU      &    8 &   16 &      69 &   89 &  100 &           93 \\
Gearbox   &    4 &    5 &      11 &   14 &   24 &           13 \\
MFPT      &   18 &   46 &      62 &   68 &   77 &           68 \\
Paderborn &   10 &   26 &      54 &   62 &   68 &           62 \\
SEU       &   20 &   23 &      87 &   96 &  100 &           95 \\
UOC       &    8 &   17 &      82 &   95 &  100 &           94 \\
XJTU      &    7 &   35 &      89 &   95 &   99 &           97 \\
\bottomrule
\end{tabular}
\caption{Tukey's five-number summary of how accurate the transformer architecture was with different hyperparameter settings. The accuracy score when using the default hyperparameter values is also shown in the right-hand column.}
\label{tab:faultformer_accuracies}
\end{table}

\subsection{Results Pt. 2 -- Focusing on the Wide-Kernel CNN Architecture}

The main body of our research is about the wide-kernel CNN architecture described in Section \ref{sec:method:widekernelarchitecture}. First, we revisit how successful networks in a grid search are when trained and tested on seven different benchmark datasets. 
This grid search is the same as was analysed before in \cite{hudson2024stay}, although here our results have a different focus. We look at the average test results when setting individual hyperparameters to specific values, and we also zoom in on the interactions between the hyperparameters that were previously identified as most important. These results help us formulate guidance on how to tune the hyperparameters of the wide-kernel CNN. Following that, we also present multiple defaults, a small selection of the most effective hyperparameter combinations to try when moving to new data. 

After revisiting the grid search applied to seven benchmark datasets we shift our attention to how well wide-kernel CNNs work on manipulated versions of the CWRU data. Deliberately manipulating the data is a means to get greater control of data properties and see how they influence which hyperparameter values are best. We maintain a special interest in the kernel size in the first layer (the `wide kernel' of the wide-kernel architecture) because this seems to be the most data-dependent hyperparameter. The other hyperparameters of the network appear fairly consistent in terms of what is the correct way to tune them, but the first layer kernel size sometimes needs to be shrunk and sometimes needs to be expanded. By first resampling and then filtering we look at whether the sampling rate and the spectral content have an impact on what kernel size is best.

\subsubsection{Summarising Grid Search Results}

Now we look at how the accuracy of the wide-kernel CNN varies as we change both the hyperparameter values and the data.
Each condition's accuracy is shown in Table \ref{tab:hyperparameters_datasets_averages}. These results are based on the same grid search used previously in \cite{hudson2024stay}, which trained and tested different combinations of hyperparameters on 7 benchmarks. We revisit the grid search in order to pull out and highlight helpful information when thinking about how to tune the hyperparameters on different datasets. 
\begin{table}[htb]
\noindent
\scriptsize
\begin{tabular}{p{1.35cm}p{0.35cm}@{\hspace{0.3cm}}|p{0.5cm}p{0.5cm}p{0.4cm}p{0.65cm}p{0.25cm}p{0.25cm}p{0.4cm}}
\toprule
Hyperparameter & Value &  CWRU &  Gearbox &  MFPT &  Paderborn &   SEU &   UOC &  XJTU \\
\midrule
Kernel size  & 16  &  25 &     78 &  47 &       74 &  87 &  32 &  64 \\
layer 1                   & 32  &  31 &     78 &  48 &       84 &  80 &  35 &  59 \\
                   & 64  &  40 &     77 &  49 &       87 &  70 &  40 &  64 \\
                   & 128 &  55 &     77 &  51 &       89 &  61 &  49 &  64 \\
                   & 256 &  69 &     77 &  53 &       89 &  53 &  58 &  70 \\
\midrule
Stride  & 4   &  41 &     77 &  48 &       79 &  69 &  35 &  54 \\
layer 1            & 8   &  44 &     80 &  50 &       86 &  70 &  45 &  66 \\
                   & 16  &  47 &     75 &  51 &       89 &  72 &  49 &  73 \\
\midrule
Filters   & 8   &  43 &     74 &  53 &       84 &  74 &  57 &  69 \\
layer 1                   & 16  &  53 &     76 &  56 &       86 &  74 &  61 &  68 \\
                   & 32  &  53 &     77 &  54 &       85 &  72 &  51 &  65 \\
                   & 64  &  45 &     77 &  49 &       85 &  70 &  37 &  63 \\
                   & 128 &  37 &     79 &  45 &       84 &  67 &  28 &  61 \\
                   & 256 &  33 &     81 &  42 &       83 &  65 &  23 &  60 \\
\midrule
Kernel size   & 3   &  43 &     78 &  50 &       85 &  71 &  43 &  67 \\
layer 2                   & 6   &  45 &     77 &  49 &       84 &  70 &  43 &  62 \\
\midrule
Filters  & 8   &  44 &     78 &  52 &       85 &  74 &  49 &  68 \\
         layer 2           & 16  &  50 &     78 &  52 &       85 &  73 &  50 &  66 \\
                   & 32  &  51 &     79 &  52 &       85 &  71 &  47 &  63 \\
                   & 64  &  47 &     77 &  50 &       85 &  69 &  42 &  62 \\
                   & 128 &  40 &     76 &  47 &       85 &  68 &  37 &  63 \\
                   & 256 &  33 &     76 &  45 &       84 &  66 &  33 &  65 \\
\midrule
Kernel size  & 3   &  45 &     80 &  50 &       84 &  71 &  43 &  65 \\
             layers 3-5       & 6   &  43 &     75 &  49 &       85 &  69 &  42 &  64 \\
\midrule
Filters   & 8   &  33 &     83 &  49 &       86 &  67 &  49 &  75 \\
         layers 3-5          & 16  &  45 &     91 &  50 &       89 &  72 &  51 &  82 \\
                   & 32  &  53 &     90 &  51 &       89 &  73 &  48 &  75 \\
                   & 64  &  52 &     81 &  50 &       88 &  73 &  41 &  64 \\
                   & 128 &  46 &     67 &  49 &       82 &  70 &  36 &  52 \\
                   & 256 &  35 &     52 &  49 &       74 &  65 &  32 &  39 \\
\midrule
Mean accuracy &  & 44   &  77 &     49 &  85 &       70 &  43 & 65 \\
\bottomrule
\end{tabular}
\caption{The average performance for different values of wide-kernel CNN hyperparameters when trained and tested on different datasets.}
\label{tab:hyperparameters_datasets_averages}
\end{table}

First, we see that the accuracy tends to be higher on some benchmark datasets than others. It seems that some tasks were more difficult or were more demanding within the constraints imposed by the grid search. Differences are not unexpected however, since each dataset is distinct in terms of recording procedure and number and type of faults. 

Perhaps more interesting are the individual accuracy scores showing what happens when we tune different hyperparameters. The kernel size in the first layer reveals the most variable results. With CWRU, Paderborn and UOC there is a strong effect whereby a longer kernel does better and the longest kernel exceeds the shortest by at least 15\% accuracy. The MFPT and XJTU benchmarks exhibit a weaker effect of roughly 6\% accuracy increase between the shortest and longest kernel. The kernel size in layer 1 barely effects the results on Gearbox data. Meanwhile, a strong contradictory effect is observed when using SEU data. The shortest kernel is much better (more than 30\% better) compared to the longest kernel size. In all, it seems that longer kernels in the first layer tend to be best but it very much depends on the fault vibration data. 

The stride in the first layer is a simpler story. In the case of Gearbox it does not affect the accuracy by more than 5\% whilst in all the other datasets it is clear that the highest value (16) is best. A greater stride reduces computation by shrinking the outputs and potentially removes redundancy by decreasing the overlap of the outputs. 

Past work also identified the number of filters in the first layer as important. Good performance is often achieved when there are 16 or 32 filters. Either 16 or 32 was the best on every dataset except XJTU  (when 8 filters was 1\% better) and Gearbox. With Gearbox, the trend is that more filters gives higher accuracy and 256 beats 8 filters by about 7\% accuracy. Although there are exceptions across the datasets, normally 16 or 32 are the safest bets when tuning this hyperparameter. 

The final `important' hyperparameter is the number of filters in layers 3 to 5. We observe that either 16 or 32 works best on all the benchmark datasets. %This conclusion at least is straightforward. 

\subsubsection{Influence of Important Hyperparameters on One Another}

We now take another deep dive into the three hyperparameters identified as important by our previous work.

\begin{figure}[htb]
    \centering
    \includegraphics[width=1.0\columnwidth]{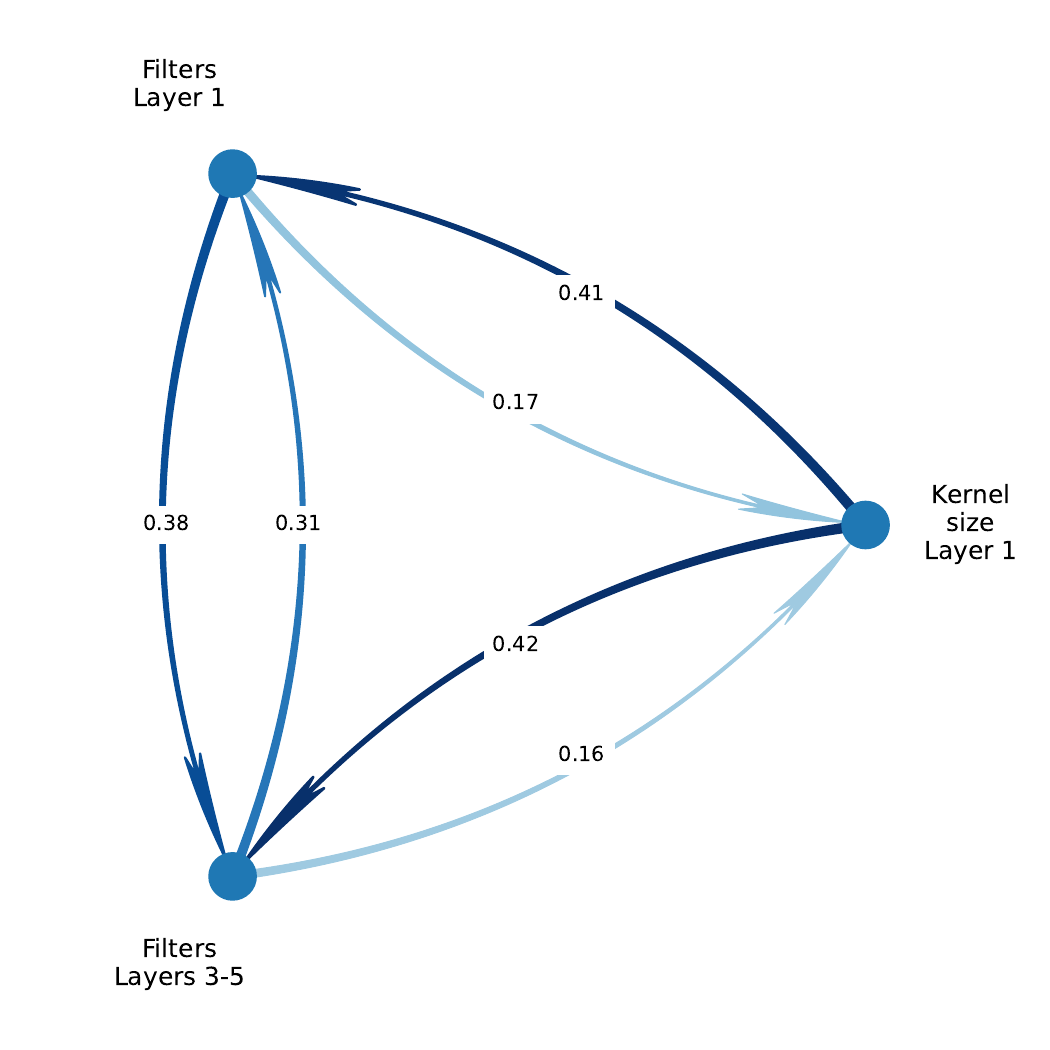} 
    \caption{The likelihood that tuning one hyperparameter (source of an arrow) will cause another (destination of an arrow) to need re-tuning.
    }
    \label{fig:influence_3_var}
\end{figure}

It is very computationally expensive to run a full grid search over all the possible hyperparameter combinations and so someone applying the network to new data might well want to know how to tune the hyperparameters one-at-a-time. 
We ask which is the best order to tune the three most important hyperparameters sequentially. The method we used in our previous paper \cite{hudson2024stay} comes in handy again here: we look at how much one hyperparameter influences another hyperparameter which has already been tuned. Because the hyperparameters are interconnected it is possible that a hyperparameter that started off being optimal no longer has the best value when a different hyperparameter changes. We look at how often this phenomenon happens and we summarise the information as an `influence' score. If A is highly influential on B then changing A most likely means changing the best value for B. We can see the numbers in Figure \ref{fig:influence_3_var}. 

The kernel size in layer 1 appears to be the most influential and therefore it appears to be the best one to tune first. Tuning it later on would likely disrupt the other hyperparameters and cause them to need re-tuning. In other words, there is often no point in tuning the number of filters in layers 1 or 3-5 beforehand, since they will need to be tuned again when the kernel size changes. After choosing the size of filters in the first layer it is best to choose how many next. The choice of how many filters in layers 3 to 5 can be delayed to the end. In brief, the best order to tune one-at-a-time is: kernel size layer 1, number of filters layer 1, number of filters layers 3-5.

\subsubsection{Multiple Defaults}

Multiple defaults provide a sequence of hyperparameter combinations to try. The goal is that by trying them out sequentially, the user will be able to identify as quickly as possible at least one hyperparameter combination that works well on their data. This provides an alternative to tuning methods that are fast but do not consider interactions between hyperparameters (tuning one-at-a-time) or are slow but explore more options (e.g., grid search, random search, Bayesian optimisation). Multiple defaults are fast to use but also flexible and able to account for interactions between hyperparameters. 

We used a brute force search to identify the global best series of hyperparameter combinations which terminated after finding 10 combinations. These are shown in Table \ref{tab:multiple_defaults} with the observed performance on the seven benchmarks. 

\begin{table*}[htb]
\centering
\small
\begin{tabular}{rrrrrrrrrrrrrrr}
\toprule
No. & KS 1 & S 1 & F 1 & KS 2 & F 2 & KS 3+ & F 3+ & CWRU & Gearbox & MFPT & Paderborn & SEU & UOC & XJTU \\
\midrule
1 & 256 & 16 & 256 & 3 & 16 & 3 & 64 & 0.96 & 0.89 & 0.93 & 0.88 & 0.57 & 0.99 & 0.96 \\
2 & 16 & 16 & 128 & 6 & 128 & 3 & 64 & 0.08 & 1.00 & 0.07 & 0.26 & 0.97 & 0.12 & 0.58 \\
3 & 256 & 8 & 32 & 3 & 32 & 6 & 64 & 0.98 & 0.68 & 0.99 & 0.98 & 0.16 & 0.98 & 0.99 \\
4 & 16 & 4 & 8 & 6 & 32 & 3 & 8 & 0.44 & 0.40 & 0.61 & 0.35 & 1.00 & 0.46 & 0.69 \\
5 & 128 & 16 & 32 & 6 & 64 & 6 & 128 & 0.97 & 0.39 & 1.00 & 1.00 & 0.35 & 0.84 & 0.49 \\
6 & 256 & 4 & 64 & 6 & 128 & 6 & 128 & 1.00 & 0.00 & 0.22 & 0.29 & 0.07 & 0.40 & 0.15 \\
7 & 256 & 4 & 256 & 6 & 8 & 3 & 16 & 0.74 & 0.94 & 0.66 & 0.95 & 0.51 & 0.60 & 1.00 \\
8 & 256 & 16 & 32 & 6 & 256 & 6 & 128 & 0.51 & 0.16 & 1.00 & 0.91 & 0.13 & 1.00 & 0.55 \\
9 & 256 & 16 & 64 & 6 & 64 & 3 & 64 & 0.96 & 0.85 & 0.97 & 0.54 & 0.25 & 1.00 & 0.43 \\
10 & 32 & 8 & 32 & 3 & 8 & 6 & 128 & 0.51 & 0.23 & 1.00 & 0.69 & 0.80 & 0.22 & 0.25 \\
\bottomrule
\end{tabular}
\caption{How multiple defaults perform on the seven benchmark datasets.}
\label{tab:multiple_defaults}
\end{table*}

We observe from Table \ref{tab:multiple_defaults} that the multiple defaults are complementary to each other, with the first entry performing relatively well on most benchmarks but having mediocre performance on the SEU dataset. The second entry picks up the slack and has very high performance on SEU. The third entry has excellent performance on the Paderborn dataset, improving on the previous two entries by 10\%. The hyperparameter values of the first three entries tell a similar story. The first entry has the largest kernel size and many filters in the first layer. By contrast, the second entry has the smallest kernel size but many more filters in the second layer. The third entry also has the largest kernel size but combines it with a modest number of filters. 

\begin{figure}[htbp]
    \centering
    \includegraphics[width=1.0\columnwidth]{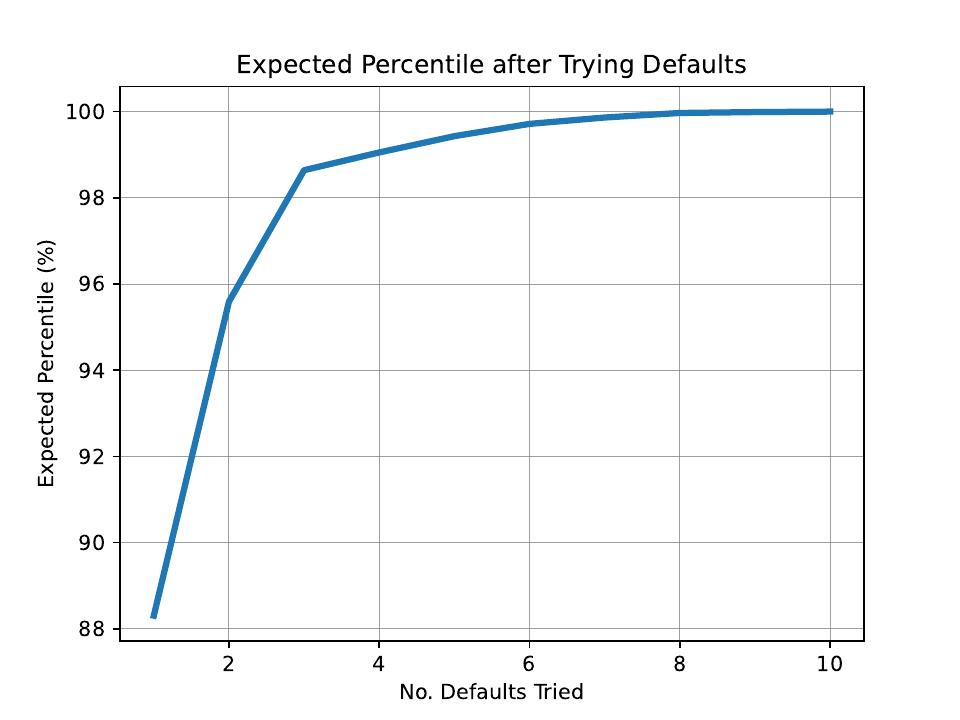} 
    \caption{Expected performance relative to other hyperparameter settings, after trying multiple defaults. Calculated by taking the best-performing CNN tried so far per benchmark, then averaging the quantile of how well those CNNs did. 
    }
    \label{fig:multiple_defaults}
\end{figure}

To illustrate how quickly these defaults identify a high-performing combination of hyperparameters, the average quantile (across benchmarks) of the best performer after trying $n$ hyperparameter combinations is visualised in Figure \ref{fig:multiple_defaults}. We see that the expected performance increases quickly and there is already very little room to improve after trying three of the defaults. This gives hope that it will be possible to rapidly identify an effective combination of hyperparameters when working on new datasets. From a user's perspective, a lot of time can be saved from the usually arduous process of hyperparameter tuning. 

To get some idea of generalisation to new datasets, we repeatedly run and evaluate the search for multiple defaults using leave-one-out validation. %This means we use six benchmarks to find defaults and then use the remaining benchmark to see how well the default values do. The process is repeated seven times so that each benchmark has a turn of being left out. 
The average best performance when generalising to an unseen benchmark is then computed. On average the multiple defaults get to the 92nd percentile on unseen data. Multiple defaults make it possible to identify hyperparameter combinations that perform well on new data, relative to the great majority of other hyperparameter combinations included in the full grid search. 

\subsubsection{Resampling}

The results of a grid search over the three most important hyperparameters, performed once each for the different resampling conditions, are brought together in the results here.
Visualising the results makes it possible to spot any obvious trends in how the (increasing levels of) resampling impact the performance of different hyperparameter settings. 
% We present several analyses on both CWRU 48 kHz data to see how much each hyperparameter seems to influence performance. These analyses include using a method based on Shapley values and using a method based on d-dimensional earth mover's scores to pinpoint important hyperparameters. 
% Here, we include the results obtained when resampling CWRU data, starting from data originally sampled at 48 kHz. 

\begin{figure}[htb]
    \centering
    \includegraphics[width=1.0\columnwidth]{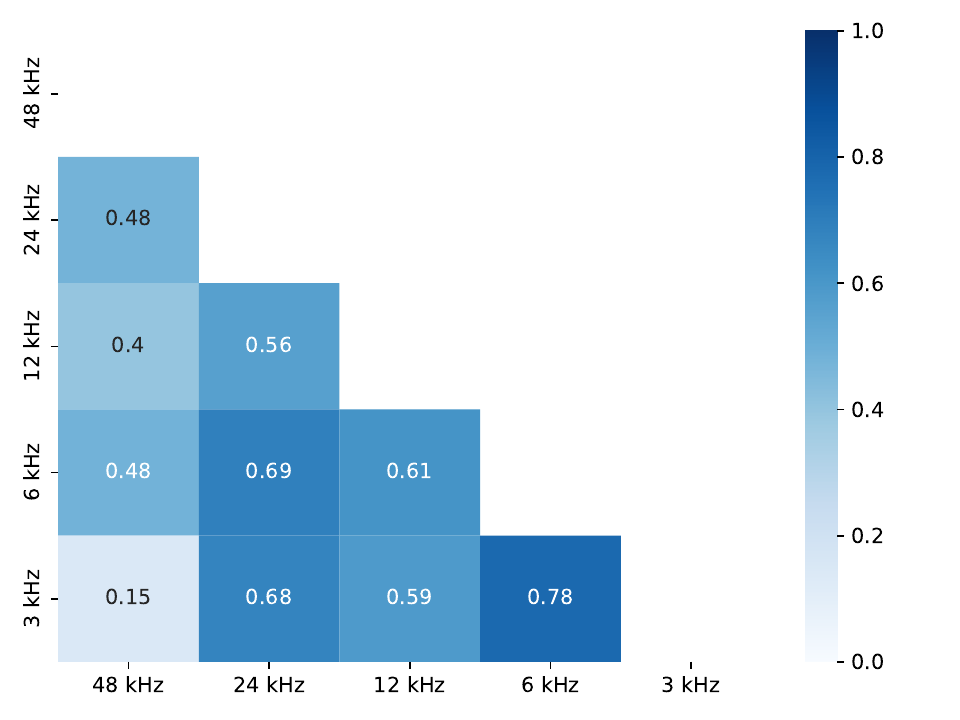} 
    \caption{Correlation between the same CNN configurations in terms of performance when using different resampling conditions, starting from 48 kHz. 
    }
    \label{fig:correlations_resampling_48kHz}
\end{figure}

First, we consider how much the accuracy scores change as a consequence of resampling. Correlations tell us whether the best-performing combinations of hyperparameters on one version of the data also remain the best after resampling, and vice versa with the worst performers. The computed correlations are shown in Figure \ref{fig:correlations_resampling_48kHz}. All correlations are positive which implies that the hyperparameter values that do well on one version of the data have a tendency to perform well on data resampled to a different rate. The weakest correlation is between the original 48 kHz data and the most resampled data of 3 kHz. This is not surprising since they are the furthest away from each other in terms of data properties.
%The weakened correlation between 48 kHz and 3 kHz gives us an initial

Before moving on we must make a brief disclaimer about the correlations, specifically there is a ceiling effect because many networks achieve close to the maximum accuracy of 1.0, and, the training and testing of neural networks is stochastic and so there may be some random noise in the accuracy scores. There is a chance that these statistical quirks have inflated or deflated the correlations, and so we do not report p-values which might be biased by these effects. %Instead, we view the results as indicative. The results give an indication that hyperparameter settings that are stronger on one version of the data are more likely to be stronger when also used on a resampled version of the data, but room is also left for differences between sampling rates to emerge. 

\begin{figure}[htbp]
    \centering
    \includegraphics[width=1.0\columnwidth]{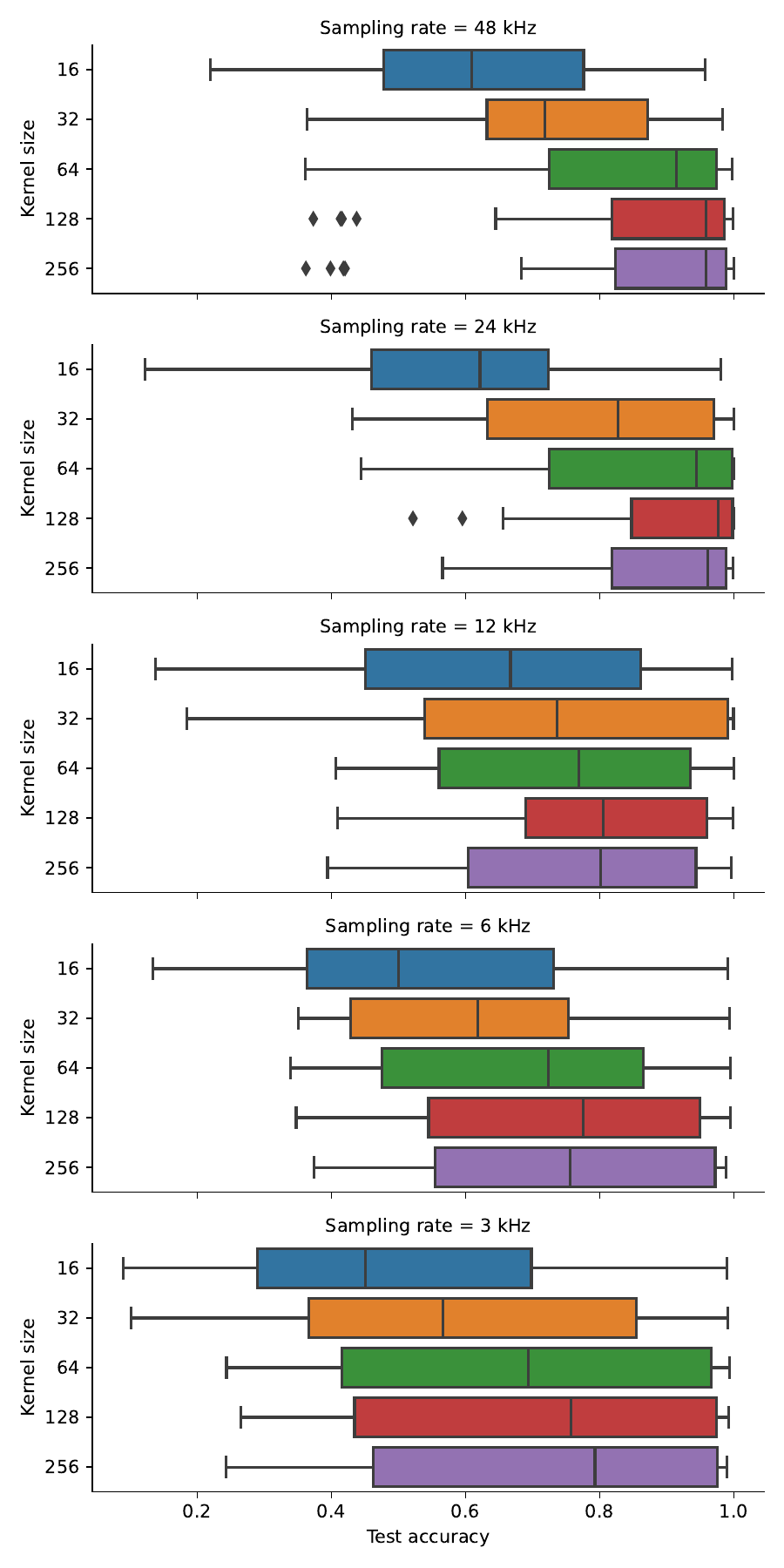} 
    \caption{Accuracy scores for different kernel sizes in the first convolutional layer, as tested on different resampling conditions for data originally sampled at 48 kHz. 
    }
    \label{fig:kernelsize_resampling_48kHz}
\end{figure}

We take a closer look at the kernel size hyperparameter in particular since we want to know if the kernel size needs to change when the sampling rate is altered. 
The box plots in Figure \ref{fig:kernelsize_resampling_48kHz} let us see more clearly which specific kernel sizes are best and not only how important the hyperparameter is overall. These plots confirm that larger kernels perform better, with 128 or 256 being the best even when resampling from 48 kHz to 3 kHz. Resampling condenses the data and makes the inputs shorter and one might expect smaller kernels to become more effective as a result of increased resampling. The results however show no such trend.
When looking at the box plots we also observe that the task becomes more difficult as the data is resampled further. Either the loss of information that comes with compression or the impact of reducing the length of inputs could explain the downwards trajectory. Considering the results in Figure \ref{fig:kernelsize_resampling_48kHz}, resampling to 24 kHz at least seems to have no negative impact and so it could be unnecessary and wasteful to record at 48 kHz when a lower sampling frequency is equally good. Further research would be needed to establish this more firmly.

\subsubsection{Filtering}

In this section, instead of resampling we look at filtering the data. Filtering modifies the information that is included in the data such that certain frequencies are removed, whilst holding constant the sampling rate and the time series length. This gives us some companion results to the resampling results we just showed. The filtered data isolates the impact of the content of the signals whereas the previous resampling also made the sequences shorter and more condensed.

When processing data sampled at 48 kHz we have nine filtering conditions: low-pass with a threshold of 12 kHz, 6 kHz, 3 kHz, 1.5 kHz, 750 Hz, 375 Hz, 187 Hz, 93 Hz and 46 Hz. 
% These can be compared side by side.
% We use the methods described earlier, namely employing Shapley values and the d-dimensional earth mover's scores, to investigate the impact of filtering.
It had been suggested that a wide kernel is beneficial primarily because it filters out high-frequency noise. The filtering experiments make it possible to verify whether this really is the benefit granted by a wide kernel. 
% Here, we include the results obtained when filtering CWRU data, applying different thresholds to decide how many frequencies to filter out. 

\begin{figure}[htb]
    \centering
    \includegraphics[width=1.0\columnwidth]{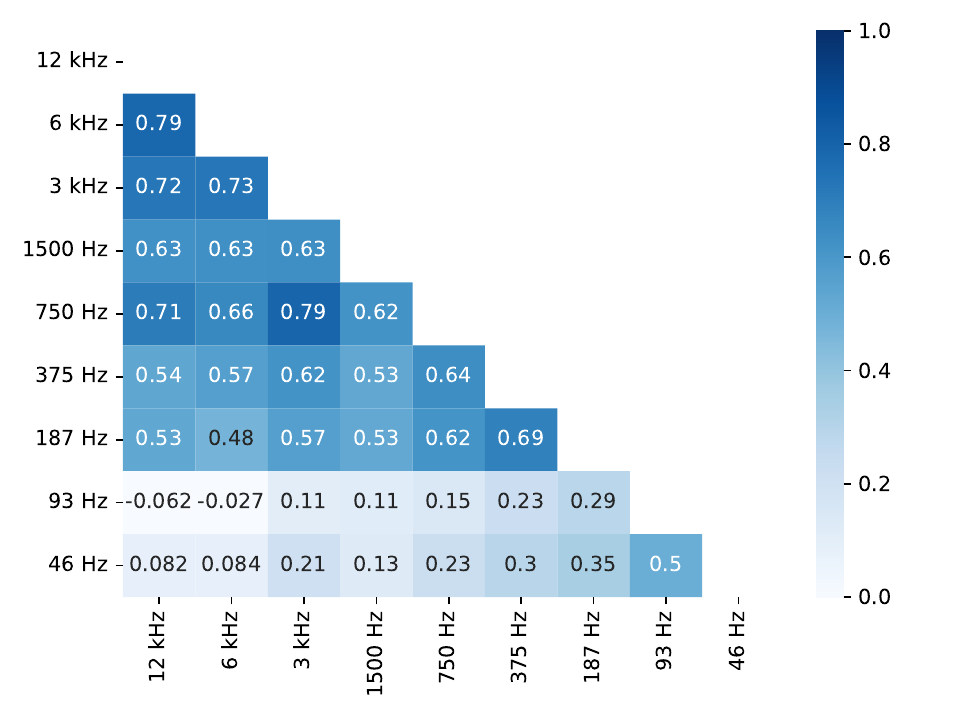} 
    \caption{Correlation between the same CNN configurations in terms of performance when using different filtering conditions. 
    }
    \label{fig:correlations_filtering_48kHz}
\end{figure}

Correlations between the different conditions appear in Figure \ref{fig:correlations_filtering_48kHz}. We see that many levels of filtering, especially the weakest forms that only remove the highest frequencies, give correlations in the range 0.5 to 0.8. These quite high correlations imply that moderate filtering did not have a large impact on how to set the hyperparameters. 
Filtering at a threshold lower than 187 Hz however causes a sudden drop in correlations. The most extreme forms of filtering evidently do affect which hyperparameter settings work well. We observe that the average accuracy is much lower when filtering at 46 or 93 Hz (the average accuracy was 0.29 in both cases) than when filtering at 3000 Hz or above (the averages ranged from 0.95 to 0.97), and there might simply be no distinction between good and bad hyperparameter settings when using extreme filtering. If the meaningful content has been filtered out of the signal, only noise will remain and the performance of different hyperparameter combinations will be essentially random (thus uncorrelated). 
%If there are no good and bad hyperparameter settings then the results will be more influenced by noise and will correlate less, which is one possible explanation for the relatively low correlations seen.
In summary, hyperparameters generalise fairly well and so the same settings do well on multiple levels of filtering, except when the strongest filtering is applied. Filtering changed the data properties but did not completely redefine which settings are good or bad.

\begin{figure}[htbp]
    \centering
    \includegraphics[width=1.0\columnwidth]{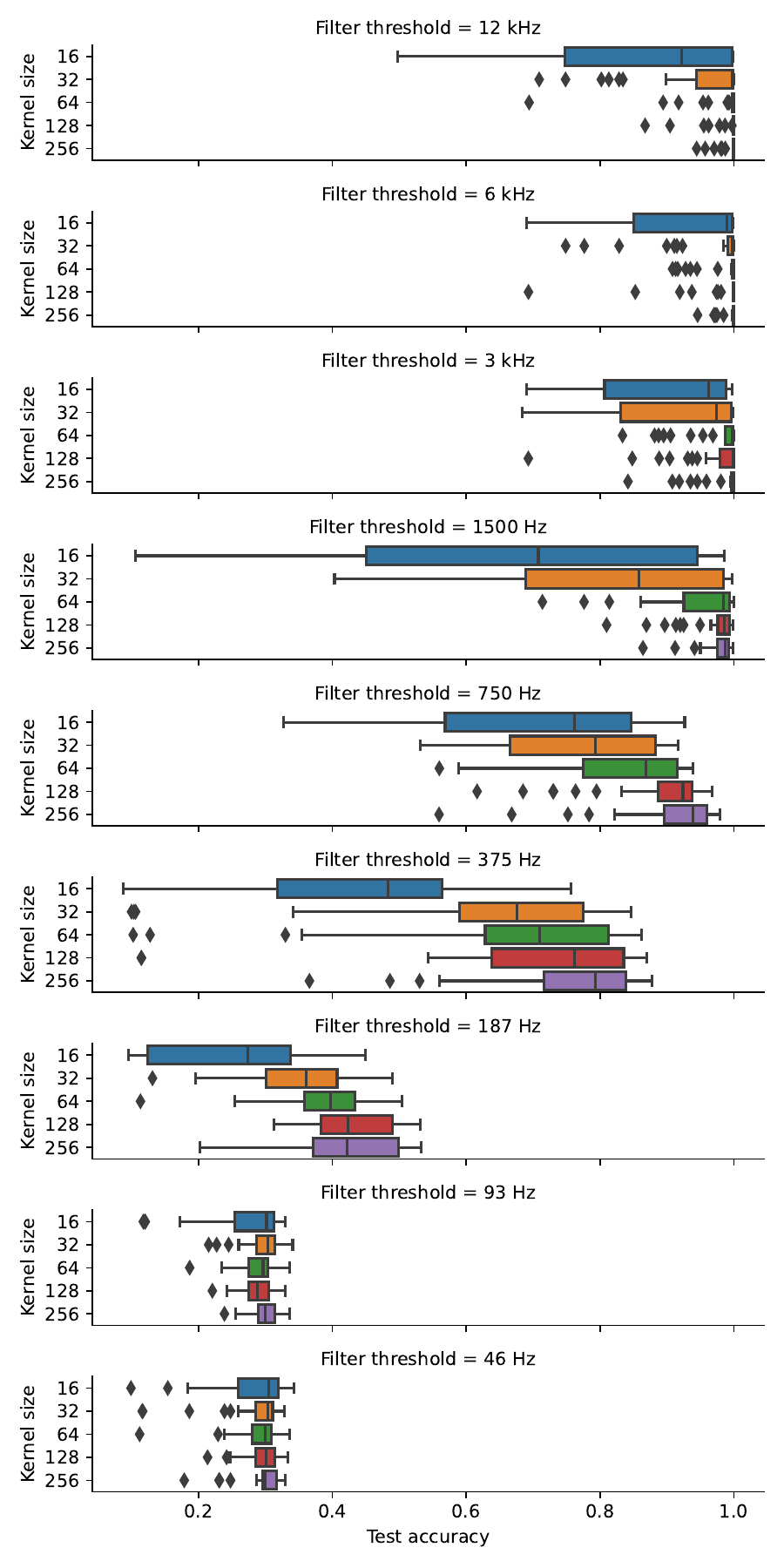} 
    \caption{Accuracy when using different kernel sizes in the first layer, as applied to different filtering conditions when using data recorded at 48 kHz. 
    }
    \label{fig:kernelsize_filtering_48kHz}
\end{figure}

In order to make a deeper dive into the kernel size in the first layer we can plot the accuracy scores visually for each kernel size. Figure \ref{fig:kernelsize_filtering_48kHz} shows the spread of accuracy scores obtained when using different lengths of kernel and when processing data with different levels of filtering. If the filtering is anything milder than 93 Hz we see that a larger kernel performs better. The value of a wide kernel does not relate to high-frequency noise as was suspected because any such noise would be removed by the filters. It seems that the benefit of a wide kernel must be explained in some other way. This is interesting because the ability to filter out high-frequency noise has been suggested as a reason why wide-kernel CNNs perform well.

\section{Conclusion} 

In our introduction, we argued that while there are many successful neural network architectures for bearing fault detection, achieving good performance in real-world applications depends on how well the hyperparameters are tuned for a specific dataset.
Benchmark datasets are useful for comparing algorithms but they might not reflect the real-world conditions where the algorithm will be applied, leading to a situation where an algorithm performs well on a benchmark but poorly on a new dataset.
Even small changes to hyperparameters can significantly impact the network's performance and yet there is often no clear guidance on how to choose the best hyperparameters in order to maximise accuracy on new data. Our research focused on explaining how hyperparameters affect accuracy on different datasets instead of searching out ever-newer neural network architectures.
%We went as far as to say that explaining how hyperparameters affect accuracy on different datasets is more important than focusing on finding ever-newer neural network architectures.% and we overall argued for a shift in focus from developing new neural network architectures to explaining how to optimise existing ones for real-world applications through better hyperparameter tuning.

Past work showed that the wide-kernel architecture we focus on in this paper is sensitive to changes in the hyperparameters. Our new explorations of a transformer-based architecture and an LSTM architecture for fault detection showed that different kinds of neural network are greatly impacted by the values of the hyperparameters, with the accuracy dropping dramatically when the hyperparameters are set poorly. This serves to emphasise the value of knowing how to set the hyperparameters soundly. Across wide-kernel the CNN, LSTM and transformer architectures we see that bad hyperparameter choices lead to low accuracy when classifying vibration data. Due to differences between datasets, we also found that no single combination of hyperparameter settings works perfectly and there is a benefit to (re)tuning the hyperparameters when moving to new data. This motivated us to identify multiple defaults for the hyperparameters of the wide-kernel CNN architecture, allowing it to be tuned very efficiently for new data. 

Keeping our focus exclusively on the wide-kernel CNN architecture, a lot of our research focused on the width of the kernel in the first layer. The kernel size is an important hyperparameter and in past literature a `wide' kernel has been considered useful for processing raw sensor data. 
Making the first layer kernel too small causes a drop to 25\% average accuracy on CWRU data from 69\% when the kernel is large. 
We had expected the value of a wide kernel to be related to either the sampling rate of the data or the presence of high-frequency noise and yet our experiments showed that neither the sampling rate nor noise is responsible for wide-kernel CNNs performing better than narrow kernel networks. 
Instead, we are left speculating about what really causes wide kernels to perform better on CWRU data. One possibility is that the wide kernel provides some sort of generic architectural advantage such as regularisation that speeds up training or makes it easier for the training process to converge to a good solution. However, if the advantage of a wide kernel is in some way generic then it is not clear why the phenomenon fails to generalise to SEU data, where a larger kernel performs worse. There could be data properties driving the choice of kernel size which we are simply unaware of. 

%After considerable exploration our investigations led to some concrete advice on how to tune the hyperparameters: most of them can be set low in order to save computation, whilst the preferred number of filters in layer 1 and in layers 3-5 seems to be 16 or 32. The kernel size in the first layer is the most sensitive to differences between datasets, and it needs to be tuned carefully. Correctly tuned hyperparameters give the final touch to a wide-kernel CNN that performs well on multiple datasets yet also is small and lightweight, making it easier to train and potentially deploy in embedded systems. If performance is lower than expected when generalising to new data then the kernel size in the first layer and the number of filters in layers 3-5 have the highest feature importance and should be tuned first. If the network is still poor-performing on new data then we also suggest an order for tuning the most important hyperparameters individually (\textit{Kernel size layer 1}, \textit{Filters layer 1} and then \textit{Filters layers 3-5}). Tuning the hyperparameters one-at-a-time saves a huge amount of computation compared to a full grid search and could be more realistic for practitioners to do. 

We of course acknowledge there are limitations to our work. 
Limitations include the fact that we have to restrict our advice only to people analysing raw vibration signals sampled at a fairly high sampling rate. Our insights do not apply to other data, such as data unrelated to (potentially faulty) rotating machines. In other words, our insights are domain-specific. Another disclaimer that should be attached to our work is that although we tried to incorporate as much data as possible, there are only a few benchmark datasets in existence. Even when using seven datasets, we do not have a statistically robust sample, so our conclusions could be influenced by sampling error. It is important to check the results on new data. 

The accuracy scores reached by neural networks are strongly influenced by how they were trained.% and so it is also important to contemplate the details of our training approach. 
One consideration is that we restricted the number of epochs to 100 when training networks and applied early stopping if a network did not improve in validation loss for more than 10 successive epochs. The training procedure must end at some point and so it is necessary to have rules for when to stop in this manner. However, a possible consequence of our stopping rules is that we do not know if networks would have improved further had they been given more time to train, although there is also no reason to assume this is true. %There is a chance that some networks which seem to perform poorly are actually effective but just relatively slow to train. At the same time, there is no reason to assume this is true and the stopping rules might have had no impact on the results. The limitation is that we cannot be completely certain. 

Turning to look back at our research as a whole, we see many smaller analyses combining to give a larger picture of the hyperparameters for one specific wide-kernel CNN architecture and also how to set them. We see the hyperparameters impacting how accurate networks are on seven different benchmark datasets. 
To future users, we present ``multiple defaults''. Users can try up to ten hyperparameter combinations in order to quickly identify a high-performing configuration of the wide-kernel CNN architecture. 

\bibliographystyle{elsarticle-num} 
\bibliography{bibliography}

\appendices

\section{Additional Details on Benchmark Datasets}
\label{sec:app:benchmarks}

In this appendix we summarise the benchmark datasets. Note that we previously used and described the datasets in \cite{hudson2024stay} and \cite{hoogen2023hyperparams}. 

\subsection{CWRU Bearing Dataset}
The CWRU bearing dataset, provided by Case Western Reserve University \cite{cwru:dataset}, represents a benchmark for fault detection experiments where various damages were gouged into bearing elements. Sensors were placed on the drive end and fan end of the machine to measure vibration signals, which were digitised into two time series and segmented into sequences of 2048 data points. The data was collected at a sampling rate of 12 kHz for both the drive-end and fan-end experiment, and the data sampled at 48 kHz for the drive-end experiment. The latter is sampled at a much higher rate and therefore contains more data points. 
For all recordings retrieved from the CWRU experiments, we extracted only the machine operation conditions with motor speed of 1797 and 1750 rotations per minute (RPM).
The damages of the bearing are inflicted at three depths (0.007, 0.014, and 0.021 inches) and across five fault locations (ball, inner race, outer race opposite, outer race orthogonal, and outer race centered) with approximately balanced samples per class except for the normal conditions. We concatenated fault conditions that were equal between the two motor speeds since they are in fact the same fault. However, in some cases, a fault did not occur in both machine operating conditions, therefore making the data somewhat imbalanced. In total for all three experiments, we identified 13 fault conditions. 

The specific choice for this dataset relates to the frequent usage within the fault detection domain ~\cite{zhang2017new,
zhang2019limited,chen2019intelligent,neupane2020bearing,liu2020personalized,piltan2020svm}, due to its public availability and its structure, which mirrors real-world industrial applications where data is typically not openly accessible. For training purposes, we split the dataset into just 20\% train and 80\% test data in order to enhance the complexity of the classification task and reduce computation time. 
For more detailed information, we refer to the original source~\cite{cwru:dataset} and the test rig displayed at \url{https://engineering.case.edu/bearingdatacenter/apparatus-and-procedures}, along with 
%Figure~\ref{picture of signals} which shows an example of the vibration signals. 

% \begin{figure}[ht]
% \includegraphics[width=1\columnwidth]{bin/example_signal.png}
% \centering
% \caption{Illustration of vibration signals retrieved from~\cite{hoogen2020improved}, representing two sensors based on the CWRU Bearing dataset. The red box highlights the duration of a single data sample (sequence).} \label{picture of signals}
% \end{figure}

\subsection{Paderborn Dataset}

The Paderborn bearing dataset serves as a benchmark for fault detection and condition monitoring of damaged rolling bearing elements, as seen in studies like~\cite{8756423,s19235300}. The dataset captures motor current signals of an electromechanical drive system and vibrations of the housing~\cite{lessmeier2016condition}. These signals are extracted using existing frequency inverters, eliminating the need for additional sensors, which was necessary in the CWRU bearing experiments. This results in more resource-efficient and cost-effective experimentation. A unique feature of the method is its ability to monitor damages in external bearings positioned within the drive system but outside the electric motor. 

In total, the data derived from the experiments represents ``healthy'', ``real damaged'', and ``artificially damaged'' bearings. The data is recorded for approximately 4 seconds at a sampling rate of 64 kHz which produces numerous data files containing around 256,000 data points each. For this study, we focused solely on the ``real damaged'' experiments, specifically targeting the ``inner race'' faults, encompassing 8 distinct conditions as highlighted by~\cite{lessmeier2016condition}. 

This resulted in 80,000 sequences (10,000 for each condition), necessitating substantial computational resources to run various CNN configurations. To manage this, we took an equally divided random sample from the dataset, retaining 10\%, or 1,000 sequences per condition. Similar to other datasets, a train/test split of 20\% and 80\% was implemented.

The Paderborn Bearing dataset, compared to the CWRU data, has fewer classes, more balanced data between classes, and a higher number of sequences, which increases network training time. Furthermore, the dataset includes three time series, two motor current signals and one vibration signal. We only used the vibration signal to align with the data types from the other datasets. 
For this experiment, we used $k$-fold cross-validation with $k=3$ to enhance the generalisability of the trained networks and tested the networks' performance with various data splits. 

\subsection{Gearbox}

The Gearbox dataset originates from the necessity to optimise and analyse industrial gearboxes, particularly in applications like wind turbines where gearbox failures can lead to significant downtime~\cite{malik2019feature}. Previous research has shown that downtime due to gearbox failures is higher compared to other components. The most common approach for examining faults in wind turbine gearboxes involves recording and analysing vibration signals, which are inherently non-stationary. The Gearbox dataset was created using the SpectraQuest Gearbox Fault Diagnostics Simulator, enabling researchers to simulate the behavior of an industrial gearbox for diagnostics and prognostics research~\cite{oedi_623}.
The dataset includes vibration signals from four sensors placed in various directions around the gearbox. The experiments were conducted under varying load conditions, ranging from 0\% to 90\%. The dataset encompasses two primary conditions: a healthy gearbox and one with a broken tooth. We combined all load variations into a binary classification task, distinguishing between healthy and broken gearboxes. This resulted in 978 sequences, each containing 2048 data points. The dataset is balanced, with 492 sequences labeled as healthy and 486 as broken.
For our study, we focused on these binary conditions and implemented a train/test split of 20/80\% to make the task more challenging and to reduce computation times. 
Additionally, the dataset allows for the separation of each operating condition combined with their respective gearbox condition, offering a nuanced perspective on the data.

This comprehensive dataset provides a robust foundation for analysing and improving fault detection in industrial gearboxes, particularly in high-stakes applications like wind turbines. The detailed setup and balanced data ensure that the classification task, although simplified by the train/test split, remains a challenging and informative endeavor for network training and evaluation.

\subsection{MFPT Dataset}

The Society for Machinery Failure Prevention Technology (MFPT) dataset \cite{SMFPT2019} is a widely used dataset for studying faults in machinery. This dataset comprises 7 outer race faults, 7 inner race faults, and a healthy baseline condition, resulting in a total of 15 unique classes. The data was collected under various load conditions to provide a comprehensive overview of fault scenarios.

Sequences of 2048 data points were created from one of the vibration sensors. Fault conditions were sampled at a frequency of 48.828 kHz, yielding 71 sequences for each fault type. In contrast, the healthy condition was sampled at a higher frequency of 97.656 kHz, resulting in an additional 858 sequences. This discrepancy in the number of sequences per condition makes the dataset inherently unbalanced.

The unbalanced nature of the dataset poses a challenge for machine learning algorithms, as it requires strategies to handle the imbalance effectively. 

This dataset provides a rich source of information for the development and evaluation of fault detection algorithms, making it an essential tool for researchers in the field of machinery failure prevention. The diversity of fault conditions and the high sampling rates ensure that the dataset can be used to test a wide range of diagnostic methods.

\subsection{XJTU}
The XJTU-SY bearing datasets are provided by the Institute of Design Science and Basic Component at Xi’an Jiaotong University (XJTU) and the Changxing Sumyoung Technology Co. The datasets contain complete run-to-failure data of 15 rolling element bearings that were acquired by conducting many accelerated degradation experiments~\cite{wang2018hybrid}. In this case, we chose to extract the last 30 minutes of every experiment containing the fault occurrence, and potentially already some vibration deviations prior to this fault. Therefore, sampling an equal length of data, containing the fault would be an appropriate strategy. Since the data are retrieved from a remaining useful life (RUL) experiment, the datasets are imbalanced in general, and also until the fault occurs, they are somewhat similar. The experiments contain two time series making it a multivariate use-case. The data are sampled at a 25.6 kHz where the sampling frame was 1.28 seconds within every minute, resulting 32,768 data points per minute. This results in 480 sequences for every of the 15 different fault conditions, totaling the dataset with 7,200 sequences. Due to the vast amount of data, we can state that this dataset is large.

\subsection{UoC}
The dataset retrieved from the University of Conneticut (UoC)~\cite{Cao2018} provides a gear fault dataset that measures vibrations with the use of accelerometers. the data are collected with a sampling frequency of 20 kHz. 
It only consists of a single sensor, therefore making it a single time series channel, i.e., univariate time series.
In total, the dataset contains eight different gear fault conditions accompanied by one healthy gear condition, resulting in a multi-class classification task with 9 unique conditions. The following conditions are gathered during the experiments; 
healthy condition, missing tooth, root crack, spalling, and chipping tip with 5 different levels of severity~\cite{cao2018preprocessing}. The dataset is balanced for all conditions. We segmented the data into sequences of 2048 data points per sequence, giving 182 sequences per condition, amounting to a fairly small dataset.

\subsection{SEU}
The Southeast University (SEU) in China has developed two sub-datasets for their gearbox datasets, both aimed at providing insights into the health of bearings and gearboxes~\cite{SEUgearboxdatasets2019}. Data collection was accomplished through a Drivetrain Dynamics Simulator, capturing eight channels of vibration data. The data encompasses five different conditions: one healthy state and four fault states, all under two operational conditions defined by rotational speed and load: 20 Hz-0 V and 30 Hz-2 V. We specifically utilised the bearing-related data, extracting three vibration channels (channels 2, 3, and 4) corresponding to the x, y, and z directions of the planetary gearbox, thus forming multivariate time series data. To increase the complexity within every fault condition, we merged the data from both rotational speeds and loads. The full recordings were employed, including the initial start-up phase. Overall, the dataset consists of 5110 sequences of 2048 data points each, which translates to 1022 sequences per class, ensuring a balanced dataset across all conditions.

\section{Additional Details on Network Architectures}
\label{sec:app:architectures}

Here we present the details of the neural network architectures explored in our work. The wide-kernel CNN architecture was previously described in several papers, including \cite{hoogen2020improved}, \cite{hoogen2023hyperparams} and \cite{hudson2024stay}.

\subsection{Long- Short-Term Memory (LSTM) Network} \label{sec:app:lstm}

\begin{figure}[ht]
    \centering
    \includegraphics[width=1\columnwidth]{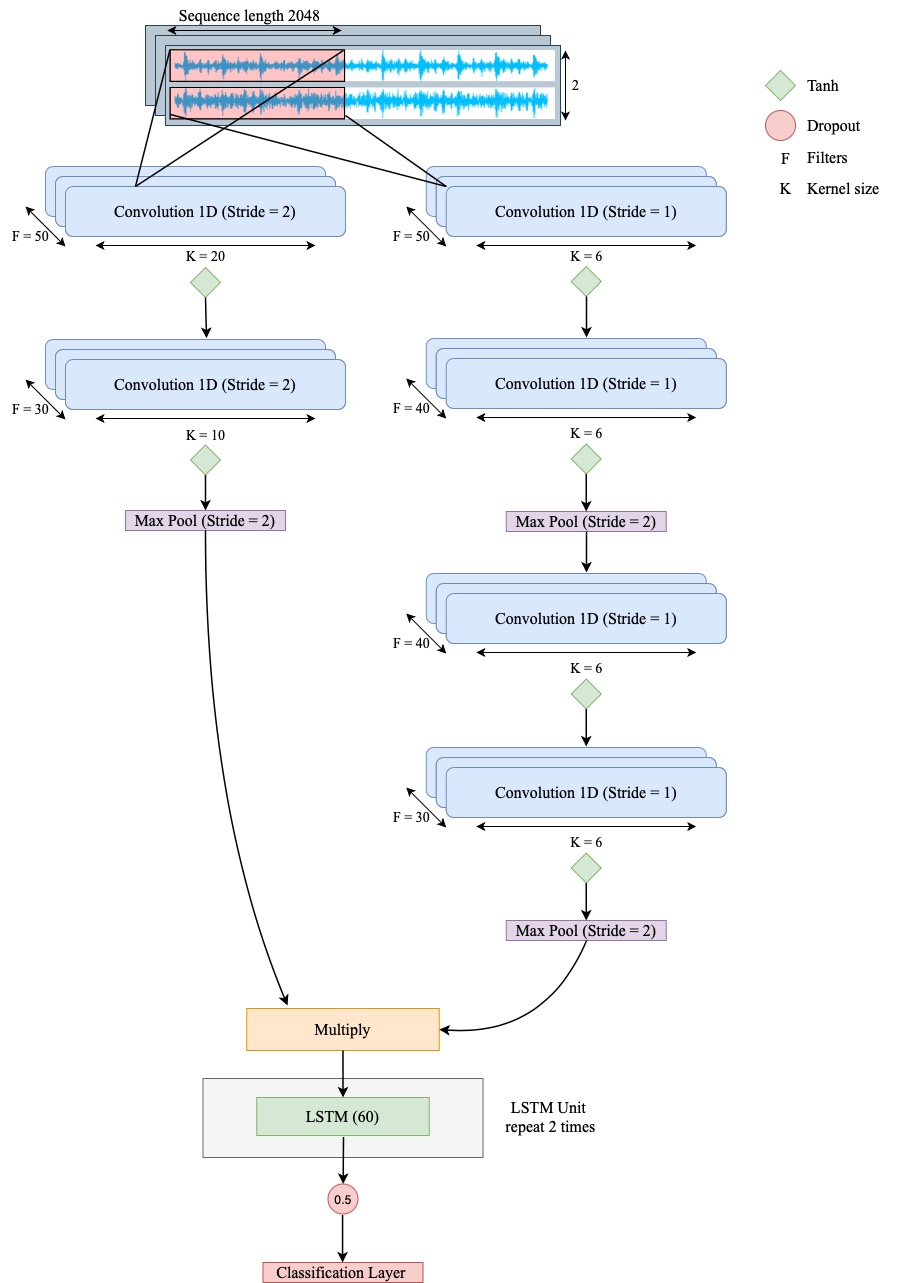} 
    \caption{Architecture of the used LSTM network, inspired by~\cite{chen2021bearing}. The network utilises two convolutional paths at different scales which are multiplied together and fed into the main LSTM. 
    }
    \label{fig:lstm_architecture}
\end{figure}

Like \cite{chen2021bearing}, we situate LSTM layers in an architecture that can be imagined as a structure with three main chambers. The first two components are CNNs which proceed in parallel and then they both pass into the third main component which is the LSTM itself. Two CNNs are used in order to achieve a ``multi-scale'' effect. A CNN containing smaller kernels gives prominence to brief or high-frequency vibrations in the signal. Features of the signal that develop more slowly are emphasised in a second CNN consisting of larger kernels. The patterns that emerge from both CNNs then proceed into the LSTM. Crucially, the patterns coming from the CNNs take on the form of a sequence, in which each step summarises a period of time from the raw signal. This is needed because the LSTM layer would be overwhelmed by the sheer number of measurement points in the raw signal without any summarisation. The LSTM component takes the shape of a pair of layers which run along the length of the input sequence and ultimately exit to a vector of numbers which captures the key details of the signal. At the end of the LSTM, fully connected layers head onwards to arrive at a final prediction of what type of fault is present in the machinery. The architecture is visualised in Figure \ref{fig:lstm_architecture}

\subsection{Transformer Architecture}
\label{sec:app:transformer}

\begin{figure}[ht!]
    \centering
    \includegraphics[width=1\columnwidth]{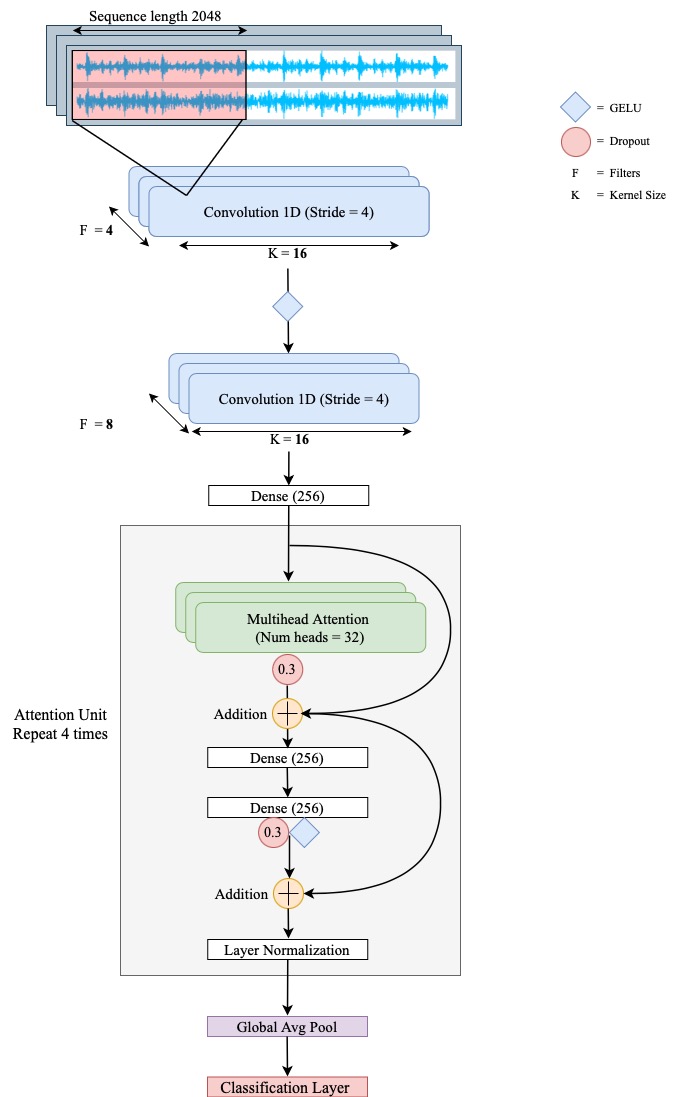} 
    \caption{Architecture of the used transformer network, inspired by~\cite{zhao2023fault}. The network utilises convolutional layers that feed into transformer layers. 
    }
    \label{fig:transformer_architecture}
\end{figure}

Much like the LSTM architecture, the transformer would also be quickly overwhelmed by the number of time steps in the raw vibration signal. Therefore, the network we use (shown in Figure \ref{fig:transformer_architecture}) receives incoming signals in a CNN component which leads to a compact sequence that then passes through to the transformer component. 
The CNN has a simple structure arranged into two layers and after passing though it the raw input signal shrinks to a sequence of roughly 1/16th the length. 
The transformer section of the architecture is larger and consists of 4 layers by default. 
When exiting the transformer layers we still have a sequence of vectors. All the pieces of the sequence need to come together into a single vector, something that is achieved using a global average pooling layer. Following the pooling layer it is possible to proceed into a single fully-connected layer which brings us to the final prediction of the network.

\subsection{Wide-Kernel Convolutional Neural Network}\label{sec:app:wk-cnn}

\begin{figure}[ht]
    \centering
    \includegraphics[width=1\columnwidth]{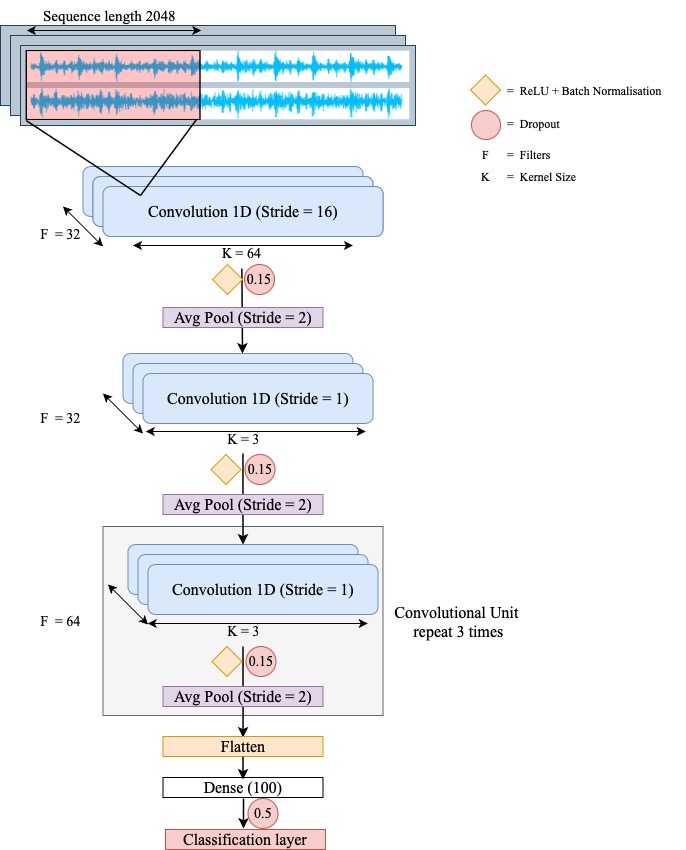} 
    \caption{Architecture of the used 1D wide-kernel CNN, inspired by~\cite{hoogen2020improvedWDCNN,hoogen2023hyperparams}. The network utilises two convolutional layers followed by a convolutional unit (in grey), which is repeated 3 times. 
    }
    \label{fig:architecture}
\end{figure}

Our network design consists of two initial convolutional layers, each employing various combinations of stride, kernel size, and filters. Subsequently, a \textbf{convolutional unit} (refer to Figure~\ref{fig:architecture}) is replicated \textit{three} times, resulting in a model with a total of five convolutional layers. In this framework, all three convolutional units share identical hyperparams, based on the rationale that deeper layers should maintain consistent characteristics for optimum learning as established in prior studies~\cite{hoogen2020improvedWDCNN,hoogen2021,zhang2017new}. Each convolutional layer is followed by local average pooling, which halves the size of the convolutional output vector by reducing the length $T$ to $\frac{T}{2}$. Finally, the reduced convolutional output is connected to two dense layers, with the last serving as the classification layer. 

Figure~\ref{fig:architecture} shows the overall network architecture with two exemplary time series as input.

In addition to the network's architecture, some additional refinements improve performance and computation time. These remained the same for every combination that was tested. 
First of all, between convolutional layers, a batch normalisation layer is added to speed up the training process. 
Second, we used the Adam stochastic optimiser.
Adam is an optimisation algorithm that leverages the strength of adaptive learning rates for each individual parameter and is well-suited for networks that analyse high dimensional input and have a lot of trainable parameters, since it is memory-light and computationally efficient. 
Furthermore, according to~\cite{kingma2014adam}, Adam is very effective with noisy signals.

\EOD

\end{document}